\documentclass[10pt,twocolumn,letterpaper]{article}

\usepackage{cvpr}              % To produce the CAMERA-READY version
\usepackage{url}
\usepackage{xspace}
\usepackage{booktabs}
\usepackage{makecell}
\usepackage{algorithm}
\usepackage{listings}
\usepackage{amsmath,amsfonts,bm}
\usepackage{algpseudocode}
\usepackage{wrapfig}
\usepackage{enumitem}
\usepackage[table]{xcolor}

\usepackage{multirow}
\usepackage{makecell}
\usepackage{pifont}

\usepackage{graphicx}
\usepackage{caption, subcaption}
\usepackage{multirow}
\usepackage{color}
\definecolor{baselinecolor}{gray}{0.9}
\usepackage{amsmath}
\usepackage{cuted}

\definecolor{cvprblue}{rgb}{0.21,0.49,0.74}
\usepackage[pagebackref,breaklinks,colorlinks,allcolors=cvprblue]{hyperref}
\usepackage[accsupp]{axessibility}  % Improves PDF readability for those with disabilities.

\newcommand{\paravspace}{\vspace{-10pt}}

\def\paperID{***} % *** Enter the Paper ID here
\title{NeAR: Coupled \underline{Ne}ural \underline{A}sset–\underline{R}enderer Stack}

\author{
Hong Li$^{1,2\ast}$ \quad
Chongjie Ye$^{3,11\ast}$ \quad
Houyuan Chen$^{4}$ \quad
Weiqing Xiao$^{5}$ \quad
Ziyang Yan$^{6}$ \\
Lixing Xiao$^{7}$ \quad
Zhaoxi Chen$^{8}$ \quad
Jianfeng Xiang$^{9}$ \quad
Shaocong Xu$^{2}$ \quad
Xuhui Liu$^{1}$ \quad
Yikai Wang$^{10}$ \\
Baochang Zhang$^{1\dagger}$ \quad
Xiaoguang Han$^{11,3}$ \quad
Jiaolong Yang \quad
Hao Zhao$^{12\dagger}$\\
\small{$^{1}$BUAA \quad}
$^{2}$BAAI \quad
$^{3}$FNii, CUHKSZ \quad
$^{4}$HKUST \quad
$^{5}$NJU \quad
$^{6}$UniTn \\
\small{$^{7}$ZJU \quad}
$^{8}$NTU \quad
$^{9}$THU \quad
$^{10}$BNU \quad
$^{11}$SSE, CUHKSZ \quad
$^{12}$AIR, THU \\
{\small
\textit{Project Page:} \href{https://near-project.github.io/}{near-project.github.io}
}
}

\begin{document}

% \twocolumn[{%
% \renewcommand\twocolumn[1][]{#1}%
% \maketitle
% \begin{center}
%     \vspace{-10pt}
%     % \includegraphics[width=\textwidth]{figures/teaser_w.pdf}
%     \includegraphics[width=\textwidth]{figures/paras1.pdf}
%     \vspace{-9pt}
%     \captionof{figure}{
%         \textbf{Comparison of NeAR and Decoupled Paradigms}.
%         Left: Visual results under target illumination. Cols. 3--5 are rendered via Blender to evaluate asset quality.
%         Insets (right of cols. 4\&5) display PBR maps (top-down: Base Color, Metallic, Roughness).
%         Baselines suffer from baked-in lighting (Trellis) or material ambiguity (HY3D-2.1).
%         Notably, HY3D-2.1 wrongly assigns high metallic values to the bread (see Metallic map, Row 1)
%         and exhibits inconsistent highlights on the robot (Row 3).
%         While our intermediate PBR decomposition (col. 5) corrects materials,
%         it struggles with complex effects like transparency (Helmet, Row 2) under standard rendering.
%         Our full Neural Renderer (col. 6) resolves this, yielding photorealistic results closest to GT.
%         Right: Quantitative results on the Glossy Synthetic dataset.
%         NeAR achieves the highest PSNR across all four tasks, demonstrating the superiority of our coupled stack.
%     }
%     \label{fig:teaser}
% \end{center}%
% }]

\twocolumn[{%
\renewcommand\twocolumn[1][]{#1}%
\maketitle
\begin{center}
    \vspace{-10pt}
    \includegraphics[width=\textwidth]{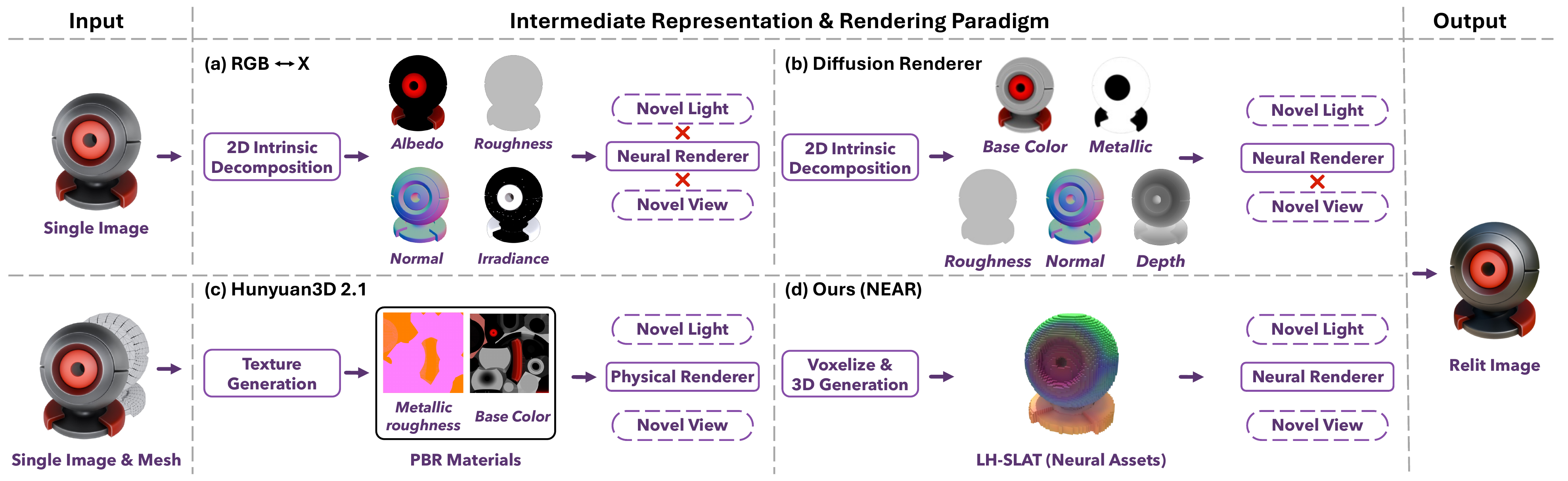}
    \vspace{-6mm}
    \captionof{figure}{\textbf{Overview of NeAR vs. Existing Single Image Relighting Frameworks.} 
    (a-b) Existing 2D methods lack explicit 3D awareness; specifically, (a) struggles to disentangle specular highlights, while both fail to guarantee multi-view consistency during relighting. 
    (c) State-of-the-art 3D generation methods decouple asset authoring from rendering, relying on ill-posed PBR decomposition that often results in material inaccuracies and baked-in artifacts. 
    In contrast, (d) \textbf{NeAR (Ours)} employs a \textbf{Coupled Neural Asset--Renderer Stack}. By utilizing the \textbf{LH-SLAT} representation, we simultaneously achieve photorealistic relighting and consistent novel-view synthesis.}
    \label{fig:paras}
\end{center}
}]

{
    \renewcommand{\thefootnote}{\fnsymbol{footnote}}
    
    \small\footnotetext[1]{Equal contribution. 
    $^{\dagger}$Corresponding authors.
    }
}

\begin{abstract}
Neural asset authoring and neural rendering have traditionally evolved as disjoint paradigms: one generates digital assets for fixed graphics pipelines, while the other maps conventional assets to images. However, treating them as independent entities limits the potential for end-to-end optimization in fidelity and consistency. In this paper, we bridge this gap with \textbf{NeAR}, a \textbf{Coupled Neural Asset--Renderer Stack}. We argue that co-designing the asset representation and the renderer creates a robust "contract" for superior generation. On the \textbf{asset} side, we introduce the \textbf{Lighting-Homogenized SLAT (LH-SLAT)}. Leveraging a rectified-flow model, NeAR lifts casually lit single images into a canonical, illumination-invariant latent space, effectively suppressing baked-in shadows and highlights. On the \textbf{renderer} side, we design a \textbf{lighting-aware neural decoder} tailored to interpret these homogenized latents. Conditioned on HDR environment maps and camera views, it synthesizes relightable 3D Gaussian splats in real-time without per-object optimization. We validate NeAR on four tasks: (1) G-buffer-based forward rendering, (2) random-lit reconstruction, (3) unknown-lit relighting, and (4) novel-view relighting. Extensive experiments demonstrate that our coupled stack outperforms state-of-the-art baselines in both quantitative metrics and perceptual quality. We hope this coupled asset-renderer perspective inspires future graphics stacks that view neural assets and renderers as co-designed components instead of independent entities.
\end{abstract}

\vspace{-4mm}    
\section{Introduction}
\label{sec:intro}

\begin{figure*}[tbp]
\small
\centering
\includegraphics[width=\linewidth]{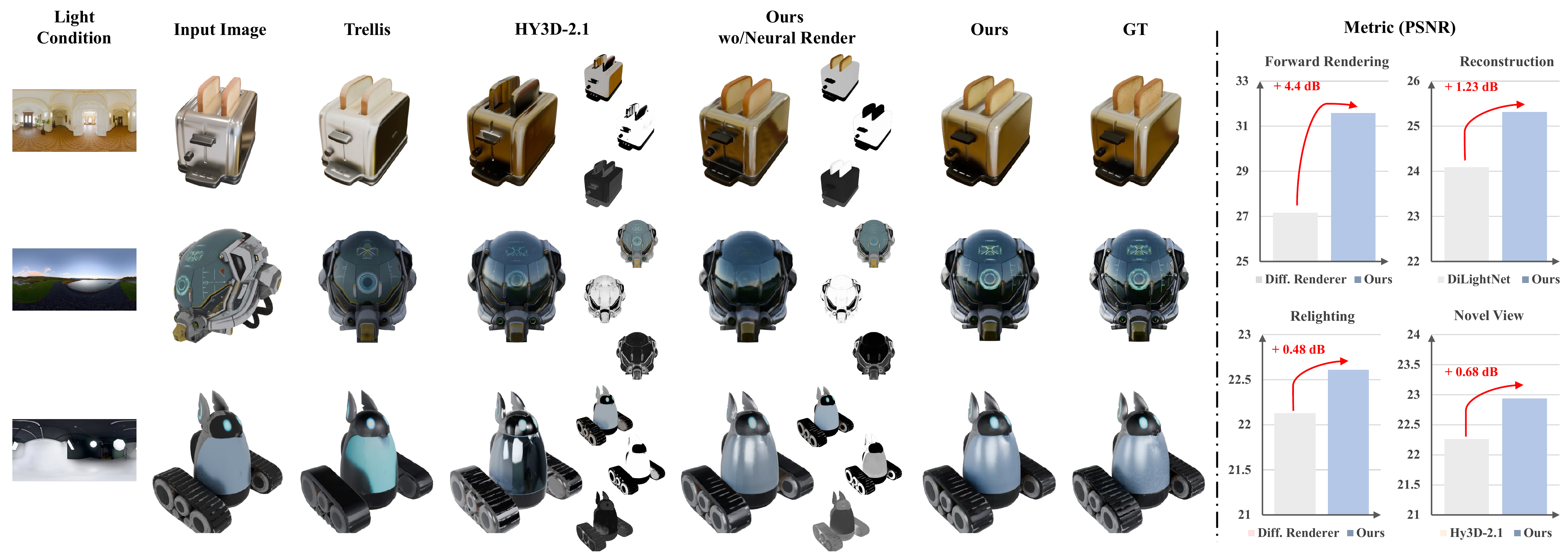}
\vspace{-6mm}
\caption{
        \textbf{Comparison of NeAR and Decoupled Paradigms}.
        Left: Visual results under target illumination. Cols. 3--5 are rendered via Blender to evaluate asset quality.
        Insets (right of cols. 4\&5) display PBR maps (top-down: Base Color, Metallic, Roughness).
        Baselines suffer from baked-in lighting (Trellis) or material ambiguity (HY3D-2.1).
        Notably, HY3D-2.1 wrongly assigns high metallic values to the bread (see Metallic map, Row 1)
        and exhibits inconsistent highlights on the robot (Row 3).
        While our intermediate PBR decomposition (col. 5) corrects materials,
        it struggles with complex effects like transparency (Helmet, Row 2) under standard rendering.
        Our full Neural Renderer (col. 6) resolves this, yielding photorealistic results closest to GT.
        Right: Quantitative results on the Glossy Synthetic dataset.
        NeAR achieves the highest PSNR across all four tasks, demonstrating the superiority of our coupled stack.
}
% \label{fig:paras}
\label{fig:teaser}
\vspace{-5mm}
\end{figure*}

Images are determined by the interaction of light with scene geometry, materials, and lighting. Classical computer graphics separates this process into asset authoring, where artists define scene properties, and rendering, where a physically based renderer simulates light transport. While effective, this separation requires substantial manual effort, computationally expensive simulations, and makes inverse reconstruction from real-world images or video challenging. Recent advances in neural graphics~\cite{zhang2024clay, wu2024direct3d, li2025triposg,zhao2025hunyuan3d,xiang2024structured,ye2025hi3dgen, sf3d, chen2025meshgen, jin2024neural, rgbx, zeng2024dilightnet, magar2025lightlab} address these limitations from two complementary directions: \emph{neural asset authoring} uses generative models~\cite{zhang2024clay, wu2024direct3d, li2025triposg,zhao2025hunyuan3d,xiang2024structured,ye2025hi3dgen, chen20243dtopia, sf3d,  chen2025meshgen} to synthesize full 3D assets for traditional pipelines, reducing manual effort, while \emph{neural renderers} map these assets—often converted into intermediate representations such as depth, normals, or shading buffers—directly to images~\cite{diffusionrenderer, rgbx, zeng2024dilightnet}, providing a data-driven alternative to analytic rendering and enabling more robust inverse inference. Fig. \ref{fig:paras} shows a comparison between our method and previous single-image relighting frameworks.

Despite recent progress in generating 3D assets with PBR materials~\cite{zhang2024clay, wu2024direct3d, li2025triposg,zhao2025hunyuan3d,xiang2024structured,ye2025hi3dgen, chen20243dtopia, sf3d,  chen2025meshgen}, a fundamental limitation remains: asset generation and neural rendering are typically developed in isolation, with assets created assuming a fixed renderer and renderers trained on static asset distributions. This separation becomes problematic when errors in asset decomposition—such as misidentified albedo or incorrect normal maps—propagate through the rendering pipeline. Because rendering is a nonlinear process, small errors in asset decomposition compound into visible artifacts like baked-in shadows or lighting inconsistencies. Fig.~\ref{fig:teaser} demonstrates this issue: existing methods rendered with traditional physically-based renderers (e.g., Blender) exhibit lighting artifacts and fail to achieve faithful relighting.

\begin{figure}[tbp]
\small
\centering
\includegraphics[width=\linewidth]{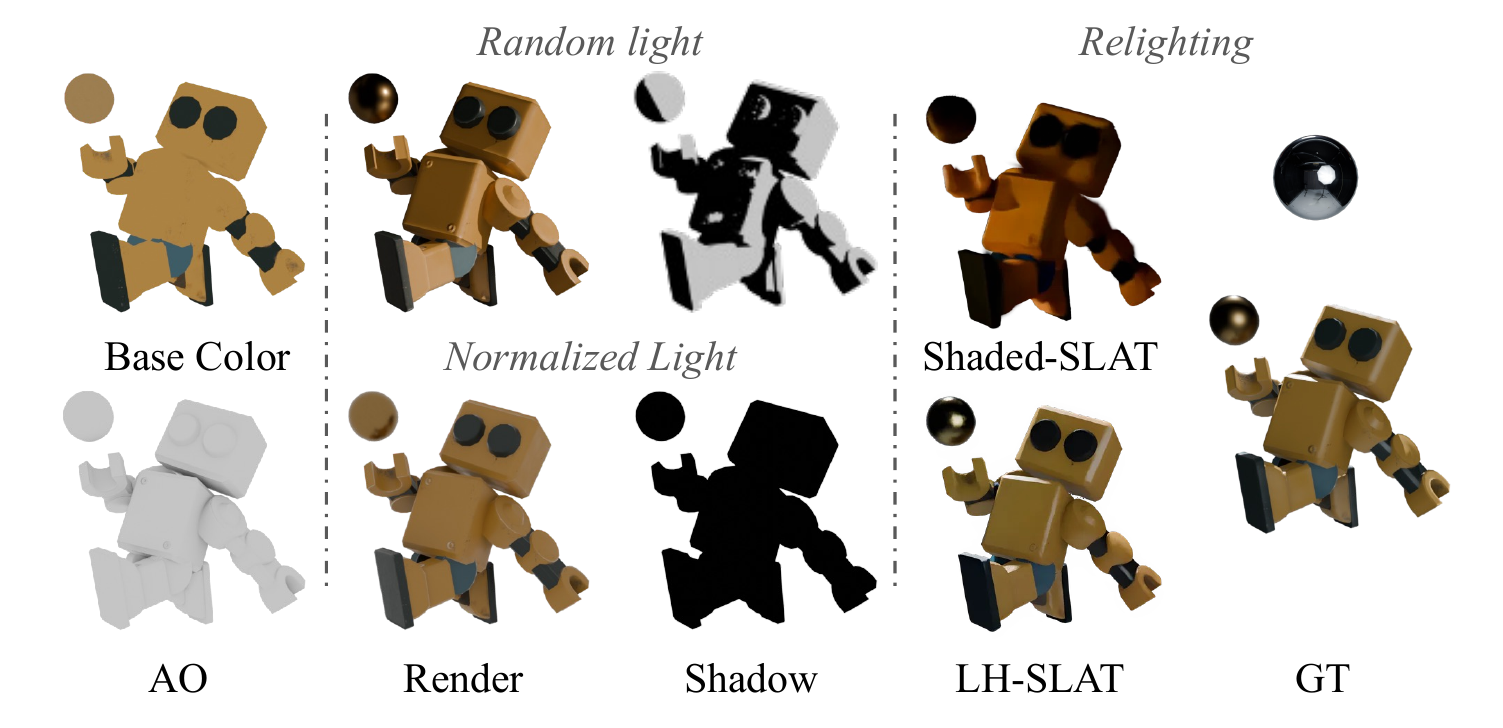}
\vspace{-6mm}
\caption{\textbf{Lighting homogenization as the bridge between assets and renderer.} 
We visualize the intrinsic components (Base Color, Ambient Occlusion), rendering results under random and uniform lighting, shadow maps, as well as relighting outputs generated respectively by Shaded SLAT and LH-SLAT. By mapping casually lit images to a canonical illumination space, LH-SLAT effectively suppresses baked-in shadows and unstable specularities while preserving geometry-consistent diffuse cues. This stable latent space serves as the robust "contract" for our lighting-aware neural renderer to enable controllable relighting.}
\label{fig:motivation}
\vspace{-4mm}
\end{figure}

To this end, we propose \textbf{NeAR}, a \emph{Coupled Neural Asset--Renderer Stack} for single-image relightable 3D generation. Our key insight is to co-design the asset representation and rendering process to enable relighting directly through a shared, lighting-homogenized latent space. On the \textbf{asset} side, we introduce a \emph{Lighting-Homogenized Structured 3D Latent (LH-SLAT)}. Unlike standard assets that rely on fragile explicit decomposition, our model lifts the casually lit input into a canonical latent form. As visualized in Fig.~\ref{fig:motivation}, this process transforms a shadow-affected representation (Shaded-SLAT) into a clean, homogenized state, effectively suppressing baked-in shadows and unstable highlights while preserving geometric cues.
On the \textbf{renderer} side, we design a \emph{lighting-aware neural renderer}. Conditioned on a lighting tokenizer, this renderer learns to interpret the homogenized latents and synthesize view-dependent appearance under arbitrary HDR environments via differentiable 3D Gaussian splatting. By unifying the representation, NeAR generates assets that naturally support real-time, high-quality relighting and novel-view synthesis with consistent materials across views.

We validate NeAR across four downstream tasks: (1) G-buffer–based forward rendering, (2) random-lit single-image reconstruction, (3) unknown-lit single-image relighting, and (4) novel-view relighting. On benchmarks including Digital Twin Category, Aria Digital Twin, and Objaverse, NeAR achieves state-of-the-art or improved performance over recent neural relighting baselines in both quantitative metrics and perceptual quality, while running at real-time frame rates without per-object optimization.

Our contributions can be summarized as follows:
\begin{enumerate}
 \item \textbf{Coupled neural asset--renderer stack.}
 We introduce NeAR, an learnable graphics stack where the neural asset representation and neural renderer are co-designed for single-image relightable 3D asset generation.

 \item \textbf{Lighting-homogenized structured neural asset.}
 We propose a Lighting-Homogenized Structured 3D Latent (LH-SLAT) that suppresses shadows and unstable highlights while preserving geometry-consistent diffuse cues in a compact, view-agnostic 3D latent.

 \item \textbf{Lighting tokenizer and lighting-aware neural 3D Gaussian renderer.}
 We design a lighting tokenizer and a lighting-aware neural 3D Gaussian renderer that map LH-SLAT, environment illumination, and view embeddings into a relightable 3D Gaussian field rendered via differentiable Gaussian splatting.

 \item \textbf{Extensive evaluation and real-time performance.}
 We demonstrate on multiple datasets and tasks that NeAR delivers state-of-the-art or better quality with strong generalization and consistent multi-view rendering, while enabling real-time feed-forward inference. 
\end{enumerate}

\section{Related Works}
\label{sec:related}

\subsection{Image relighting and inverse rendering}
Image relighting and inverse rendering lie at the intersection of geometry, material estimation, and light transport, and have been studied from both physics- and data-driven perspectives \cite{jin2024neural}. Classical methods (e.g., SIRFS) recover interpretable PBR maps (albedo, roughness, normals) via optimization with hand-crafted priors \cite{barron2013intrinsic}. While interpretable and editable, these approaches are highly ill-posed in real scenes: shadows, inter-reflections, and view-dependent highlights bias material estimation, leading to baked-in artifacts under re-rendering.

% Recent learning-based approaches generally fall into two categories. The first focuses on physically-structured decomposition~\cite{zhao2025hunyuan3d, diffusionrenderer, engelhardt2025svim3d}. Although decomposition yields interpretable assets, accurate regression from casual single-view inputs often necessitates multi-view data or costly per-object optimization to resolve ambiguities. The second category targets direct, diffusion-based 2D relighting~\cite{fortier2024spotlight, zeng2024dilightnet, zhang2025scaling, magar2025lightlab}. Methods like DiLightNet and IC-Light exploit diffusion priors to produce high-fidelity relit images with fine-grained control. However, these approaches are typically computationally expensive, stochastic, and operate in 2D, failing to guarantee multi-view consistency for 3D applications. 
Recent learning-based approaches fall into two categories. The first focuses on physically structured decomposition~\cite{zhao2025hunyuan3d, diffusionrenderer, engelhardt2025svim3d}, which yields interpretable assets but often requires multi-view data or costly per-object optimization to resolve ambiguities. The second targets diffusion-based 2D relighting~\cite{fortier2024spotlight, zeng2024dilightnet, zhang2025scaling, magar2025lightlab}. Methods such as DiLightNet and IC-Light leverage diffusion priors for high-fidelity relighting with fine control, but are computationally expensive, stochastic, and limited to 2D, lacking multi-view consistency for 3D applications.

% In this work we take a middle path: instead of directly solving a brittle PBR inversion or relying on black-box diffusion sampling, we first homogenize the input illumination into a canonical representation (LH-SLAT) and then synthesize a relightable 3D field in a feed-forward manner. This homogenize-then-synthesize strategy stabilizes downstream decoding and improves controllability.
We take a middle path: rather than brittle PBR inversion or black-box diffusion, we homogenize illumination into a canonical form (LH-SLAT) and synthesize a relightable 3D field feed-forward, improving stability and controllability.

% \subsection{Diffusion priors and 3D content generation}
% Diffusion priors and score-distillation techniques have catalyzed rapid progress in 3D synthesis from 2D models \cite{poole2022dreamfusion, tang2023dreamgaussian, yan2025learning, shi2023mvdream}. DreamFusion and follow-up works transfer 2D generative knowledge to 3D via SDS, improving fidelity at the cost of expensive iterative optimization. More recent efforts push for faster or feed-forward 3D reconstruction and native 3D generation by training on 3D datasets, or by designing architectures that decouple geometry and appearance \cite{hong2023lrm, xiang2024structured, zhang2024clay}. These native or decoder-first approaches generally offer better cross-view consistency and faster inference than optimization-based SDS pipelines, although aligning geometry and high-fidelity appearance remains challenging. Our method builds on this approach by adopting a structured 3D latent representation and a decoder that directly predicts a relightable 3D Gaussian Splatting (3DGS), replacing costly optimization with real-time, multi-view-consistent synthesis.

\subsection{Generative 3D Priors and Representations}
Diffusion priors and score-distillation sampling (SDS) have catalyzed rapid progress in text-to-3D and image-to-3D generation~\cite{poole2022dreamfusion, tang2023dreamgaussian, yan2025learning, shi2023mvdream, zhang2023temo, zhang2025ar, zhang2025tar3d}. While SDS-based methods transfer 2D generative knowledge to 3D effectively, they suffer from slow iterative optimization. Consequently, recent works have shifted toward feed-forward 3D reconstruction models trained on large-scale 3D datasets~\cite{hong2023lrm, xiang2024structured, zhang2024clay}. Specifically, Trellis~\cite{xiang2024structured} utilizes Structured 3D Latents (SLAT) to compress complex geometry and appearance into sparse tokens, enabling efficient decoding.

Concurrently, 3D Gaussian Splatting (3DGS)~\cite{kerbl20233d} has emerged as a rasterization-friendly representation supporting real-time differentiable rendering. While current feed-forward models (like LRM or Trellis) excel at geometry, they typically bake lighting into the texture, limiting downstream utility. Our method builds upon the efficiency of SLAT and 3DGS but fundamentally redesigns the generation process. We introduce a \emph{lighting-homogenized} variant of SLAT and a custom neural decoder, replacing static texture prediction with a relightable neural field.

% 
% \subsection{Structured 3D latents and 3D Gaussian Splatting}
% Structured latent representations (SLAT) compress surface-adjacent regions into sparse token sets and have shown strong inductive bias for controllable 3D decoding \cite{xiang2024structured}. Concurrently, 3D Gaussian Splatting (3DGS) has emerged as an efficient, rasterization-friendly scene representation that supports differentiable and real-time rendering \cite{kerbl20233d, qin2024instant}. Combining SLAT-style latents with 3DGS decoding lets a compact feed-forward model generate geometry, material-like attributes, and view-dependent effects with low run-time cost. Our RelitTrellis instantiates this combination: a rectified-flow-based LH‑SLAT extractor produces illumination-homogenized latents, and a lightweight decoder (IAD + EAR) synthesizes a relightable 3DGS conditioned on target lighting and view (Sec.~3).

\subsection{Relightable 3D asset synthesis}
Producing relightable 3D assets requires models to represent both intrinsic surface properties and lighting-dependent transport (shadows, speculars, interreflections). Prior works condition NeRFs, Gaussian splats or meshes on lighting inputs to enable relighting-aware outputs \cite{zeng2023relighting, jin2024neural, remondino2023critical,li2023relit, gao2024relightable, yan20243dsceneeditor, bi2024gs3, zhao2024illuminerf, Poirier_Ginter_2024}. Many approaches either use volumetric neural renderers that are costly at inference, or attempt to estimate PBR maps without lighting supervision, which leads to poor disentanglement \cite{qiu2024richdreamer, liu2024unidream, shim2024mvlight}. Some models explore large inverse-rendering architectures to predict PBR properties from sparse views, but computational cost and optimization per-object remain bottlenecks \cite{li2025lirm, zhang2024relitlrm}. Recent works \cite{engelhardt2025svim3d, tang2025rogr} employ diffusion models to generate multi-view material maps or multi-view relighted images, followed by 3D reconstruction. However, the absence of explicit 3D constraints in the generation stage makes it difficult to guarantee consistency across views.

% In contrast, our homogenize‑then‑synthesize pipeline explicitly removes unstable, scene-specific illumination before decoding. This reduces the ill‑posedness of PBR inversion and enables a single feed-forward decoder to produce relightable 3DGS with real-time rendering and improved multi-view consistency (see Sec.~1 and Sec.~3). The design occupies an intermediate point between interpretable PBR pipelines and powerful but costly diffusion-based renderers, combining stability, controllability, and practical speed for relightable asset creation.

% In contrast, our \emph{homogenize-then-synthesize} pipeline explicitly removes unstable, scene-specific illumination before decoding. This reduces the ill-posedness of PBR inversion and enables a single feed-forward decoder to produce relightable 3DGS with real-time rendering and improved multi-view consistency. By coupling asset homogenization with a lighting-aware renderer, NeAR achieves both the stability of interpretable pipelines and the fidelity of neural rendering.

In contrast, our \emph{homogenize-then-synthesize} strategy pipeline explicitly removes unstable, scene-specific illumination before decoding. This mitigates ill-posed PBR inversion, enabling a feed-forward decoder to produce relightable 3DGS with real-time consistency. NeAR thus combines the stability of interpretable pipelines with the fidelity of neural rendering.

\begin{figure*}[thb] 
    \centering
    \includegraphics[width=\textwidth]{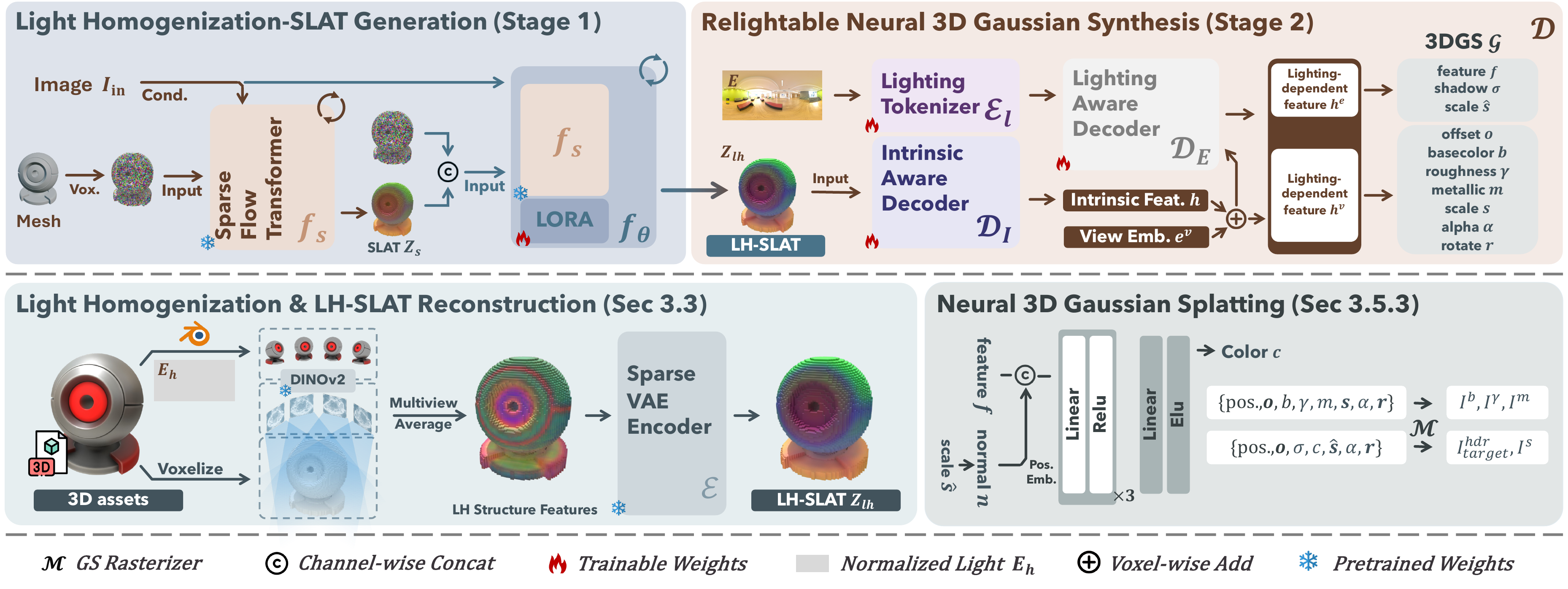}
    \vspace{-5mm}
    \caption{\textbf{Pipeline of NeAR as a coupled neural asset-renderer stack.} 
    \textbf{Top (Inference Stage):} An end-to-end inference pipeline. Given a single image and a \textbf{geometry prior} (e.g., mesh from HY3D), Stage 1 utilizes a rectified-flow backbone with LoRA adaptation to predict the \textbf{Lighting-Homogenized SLAT (LH-SLAT)}. This latent acts as a bridge, which is then consumed by the Stage 2 lighting-aware neural renderer to synthesize relightable 3DGS under novel illumination and viewpoints.
    \textbf{Bottom-Left (Data Prep):} Offline construction of ground-truth LH-SLATs by rendering assets under homogenized illumination and encoding them via a sparse VAE.
    \textbf{Bottom-Right (GS Decoding \& Rendering):} Detailed architecture of the 3DGS decoding head, which predicts Gaussian attributes from lighting-dependent features, followed by a differentiable rasterizer $\mathcal{M}$ that renders the final HDR image, shadow and PBR auxiliary maps.}
    \label{fig:pipeline}
    \vspace{-5mm}
\end{figure*}

\section{Method}
\label{sec:method}
\subsection{Preliminary}
% \textbf{3D Gaussian Splatting (3DGS).} 
% 3DGS~\cite{kerbl20233d} represents a breakthrough in neural rendering by employing anisotropic 3D Gaussians as explicit scene representations. 
% Each Gaussian is parameterized by its center $\boldsymbol{x} \in \mathbb{R}^3$, opacity $\sigma \in [0,1]$, and covariance $\Sigma \in \mathbb{R}^{3 \times 3}$, which is decomposed into a rotation quaternion $\boldsymbol{r}$ and scaling vector $\boldsymbol{s}$:
% \begin{equation}
% \Sigma = R S S^T R^T .
% \end{equation}
% For rendering, Gaussians are projected to 2D via the covariance transformation:
% \begin{equation}
% \Sigma^{\prime} = J V \Sigma V^T J^T ,
% \end{equation}
% where $J$ is the Jacobian of projection and $V$ is the view matrix. 
% Pixel color is computed via alpha blending:
% \begin{equation}
% C(\mathbf{p}) = \sum_{i \in N} T_i \alpha_i c_i, \quad
% \alpha_i = \sigma_i e^{-\frac{1}{2} (\mathbf{p} - \mu_i)^T {\Sigma'}^{-1} (\mathbf{p} - \mu_i)} ,
% \end{equation}
% where $\mathbf{p}$ is the pixel coordinate, $\boldsymbol{\mu}_i$ is the projected Gaussian center, $N$ is the ordered list of Gaussians intersecting the ray, and $T_i = \prod_{j=1}^{i-1}(1 - \alpha_j)$ is the transmittance. 
% This formulation enables differentiable, real-time rendering.

\textbf{3D Gaussian Splatting (3DGS).} 
3DGS~\cite{kerbl20233d} represents scenes with anisotropic Gaussians, rendered via splatting and $\alpha$-blending: $C = \sum_{i \in \mathcal{N}} c_i \alpha_i \prod_{j<i}(1 - \alpha_j)$.
Crucially, standard 3DGS models color $c_i$ using Spherical Harmonics (SH), which inherently bakes static lighting into the representation. To enable relighting, we forego SH and predict color dynamically conditioned on target illumination.

\textbf{Structured 3D Latents (SLAT).} 
Following Trellis~\cite{xiang2024structured}, we use SLAT to encode 3D assets efficiently.
A SLAT $\mathcal{Z} = \{(\boldsymbol{z}_k, \boldsymbol{p}_k)\}_{k=1}^{K}$ consists of $K$ active feature tokens, where each token $\boldsymbol{z}_k \in \mathbb{R}^D$ is associated with a coordinate $\boldsymbol{p}_k$ in a sparse voxel grid. 
This representation focuses capacity on surface regions ($K \ll N^3$) and supports diverse decoding heads.
However, standard SLATs blindly encode input appearance—including shadows and highlights. 
Our goal is to transform $\mathcal{Z}$ into a \emph{lighting-homogenized} form, canonicalizing the appearance to a uniform illumination while preserving geometry.

\subsection{Overview of NeAR}
\label{sec:overview}
The challenge in single-image 3D relighting lies in disentangling lighting from intrinsic object properties, since shadows, highlights, and interreflections are inherently entangled with geometry. 
To avoid unstable PBR inversion and black-box neural generation, we propose a \emph{homogenize-then-synthesize} framework that functions as a coupled stack.
NeAR first extracts a Lighting-Homogenized SLAT (LH-SLAT) from the input image to neutralize lighting effects, then decodes a relightable 3DGS.  
Our framework consists of two stages:

% \textbf{Stage~1: Light Homogenization-SLAT Generation.}  
% % A rectified flow model $f_\theta$ maps the casually lit input image $I_{\text{in}}$ into a lighting-homogenized latent $Z_{\text{lh}}$: 
% A rectified flow model $f_s$ and its Lora variant $f_\theta$ map the unknown-lit input image $I_{\text{in}}$ to a lighting-homogenized latent $Z_{\text{lh}}$.: 
% \begin{equation}
% Z_{\text{lh}} = f_{\theta}(z_s, I_\text{in})= f_{\theta}(f_{s}(I_{\text{in}}), I_\text{in}).
% \label{eq:stage1}
% \end{equation}
% $Z_s$ represents the shaded SLAT inferred directly from $I_\text{in}$, $Z_{\text{lh}}$ suppresses unstable shadows and highlights while preserving geometry-consistent cues.

% \textbf{Stage~1: Lighting Homogenization.}  
\textbf{Stage~1: Light Homogenization-SLAT Generation.}
We first utilize the pre-trained flow model $f_s$ to map the arbitrarily lit input $I_{\text{in}}$ into an initial shaded SLAT $Z_s$.
Operating within this sparse voxel space, we employ a LoRA-adapted model $f_\theta$ to steer the latent representation from $Z_s$ toward a Lighting-Homogenized SLAT (LH-SLAT) $Z_{\text{lh}}$:
\begin{equation}
Z_{\text{lh}} = f_{\theta}(Z_s, I_\text{in}) = f_{\theta}(f_{s}(I_{\text{in}}), I_\text{in}).
\label{eq:stage1}
\end{equation}
Specifically, $Z_{\text{lh}}$ suppresses the baked-in shadows and highlights inherent in $Z_s$, establishing a stable light-homogenized space. 
This representation preserves essential geometry-material-light interactions, yielding a unified and generalizable foundation for the relighting task.

\textbf{Stage~2: Relightable Neural 3DGS Synthesis.}  
Leveraging the homogenized representation $Z_{\text{lh}}$, a feed-forward decoder $\mathcal{D}$ synthesizes a relightable Gaussian field $\mathcal{G}$. 
This process is conditioned on the target view $\mathbf{v}_{\text{target}}$ and the target illumination $L_{\text{target}}$, encoded via $\mathcal{E}_l$:
\begin{equation}
\mathcal{G} = \mathcal{D}\big(Z_{\text{lh}}, \mathbf{v}_{\text{target}}, \mathcal{E}_l(L_{\text{target}})\big).
\label{eq:stage2}
\end{equation}
Finally, the relighted image is rendered using a differentiable GS rasterizer $\mathcal{M}$:
\begin{equation}
I_{\text{target}} = \mathcal{M}(\mathcal{G}, \mathbf{v}_{\text{target}}).
\label{eq:render}
\end{equation}

In the following, we describe Stage~1 (Sec.~\ref{sec:stage1}) and Stage~2 (Sec.~\ref{sec:stage2}) in detail.

\subsection{Light Homogenization \& LH-SLAT Rec.}
\label{sec:stage1}
The first stage generates a Lighting-Homogenized Structured 3D Latent (LH-SLAT) $Z_{\text{lh}}$ from a single input image $I_{\text{in}}$. This representation serves as a stable, illumination-invariant substrate for downstream synthesis.

\textbf{Lighting Homogenization.}  
% We define the homogenized light as a uniform, white ambient environment illumination. This eliminates hard shadows and promotes a more uniform distribution of diffuse and specular reflection energy (Fig.~\ref{fig:motivation}). We extract SLAT features under this lighting to create an intermediate representation suitable for relighting.
We define the homogenized light $E_h$ as a uniform, white ambient environment illumination. 
Extracting SLAT features under such lighting captures intrinsic geometric and material cues uncorrupted by transient lighting effects, serving as a robust basis for relighting.

\textbf{LH-SLAT Reconstruction.}  
To train $f_{\theta}$, we prepare paired data $(I_{\text{in}}, Z_{\text{lh}})$ via multi-step rendering of 3D assets under homogenized lighting. As shown in Fig.~\ref{fig:pipeline} top left corner, we first generate the ground-truth homogenized latents $Z_{\text{lh}}$: (1) for each 3D asset, we render $N$ views under our defined homogenized illumination; (2) we extract dense 2D visual features using a pre-trained DINOv2 model; (3) these features are back-projected into a sparse 3D voxel grid; (4) finally, this sparse grid is compressed by a pre-trained SLAT VAE encoder to obtain $Z_{\text{lh}}$.  Second, to create the corresponding input $I_{\text{in}}$, we render $M$ additional images of the same asset under diverse, random lighting conditions and camera poses.

Optionally, for highly reflective materials, we extract Basecolor SLAT $Z_{\text{bc}}$ from multi-view basecolor renderings, concatenating with $Z_{\text{lh}}$ to retain base color information.
% Optionally, for high-reflectance materials that may lose base color information under uniform illumination, we extract basecolor SLATs $z_{\text{bc}}$ from multi-view basecolor renderings using the same feature extraction and encoding process as for $z_{\text{ln}}$, as a complement to $z_{\text{ln}}$.

% \textbf{LH-SLAT Generation.}  
\subsection{LH-SLAT Generation}  
As shown in Fig.~\ref{fig:pipeline} top right corner, we use a rectified flow model $f_{\theta}$ to generate the lighting-homogenized SLAT $Z_{\text{lh}}$ from the input image $I_{\text{in}}$. The rectified flow model is trained to learn the mapping between the arbitrarily lit image and the corresponding latent representation under our homogenized lighting conditions. 
Specifically, we utilize a pre-trained SLAT rectified flow model $f_{s}$ to generate the shadowed SLAT $Z_{s}$ from the input image $I_{\text{in}}$, and subsequently fine-tune $f_{s}$ using LoRA ~\cite{hu2022lora} in the sparse voxel space ~\cite{xiang2024structured} to achieve lighting homogenization.
The loss function for training is the conditional flow matching loss $\mathcal{L}_{stage1}$:
\begin{equation}
\mathcal{L}_{stage1}
=\mathbb{E}_{t,\boldsymbol{z}_0,\boldsymbol{\epsilon}}\|\boldsymbol{v}_\theta(\boldsymbol{z}, Z_s, I_{in}, t)-(\boldsymbol{\epsilon}-\boldsymbol{z}_0)\|^2_2,
\label{eq:lora}
\end{equation}
where $\boldsymbol{z}(t)=(1-t)\boldsymbol{z}_0+t\boldsymbol{\epsilon}$ is the linear interpolation between the data sample $\boldsymbol{z}_0$ and noise $\boldsymbol{\epsilon}$, and $\boldsymbol{v}_\theta$ approximates the time-dependent vector field.  If the optional basecolor SLAT $\boldsymbol{z}_{\text{bc}}$ is used, it is concatenated with $\boldsymbol{z}_{\text{lh}}$ to provide additional color information to the subsequent stage.

\begin{table*}[htbp]
    \centering
    \small\caption{Quantitative comparison against state-of-the-art methods across four sub-tasks.}
    \label{tab:comp}
    \vspace{-2mm}
    \resizebox{2\columnwidth}{!}{
    \begin{tabular}{lcccccccccccc}
    \toprule
    \multirow{2}{*}{\quad} & \multicolumn{3}{c}{\textbf{ADT \citep{pan2023aria}}} &
    \multicolumn{3}{c}{\textbf{DTC \citep{dong2025digital}}} &
    \multicolumn{3}{c}{\textbf{Objaverse data \citep{deitke2023objaverse}}} &
    \multicolumn{3}{c}{\textbf{Glossy Synthetic dataset \citep{nero}}} \\ 
    \cmidrule(lr){2-4} \cmidrule(lr){5-7} \cmidrule(lr){8-10} \cmidrule(lr){11-13}  
    & LPIPS$\downarrow$ & PSNR$\uparrow$ & SSIM$\uparrow$ & LPIPS$\downarrow$ & PSNR$\uparrow$ & SSIM$\uparrow$ & LPIPS$\downarrow$ & PSNR$\uparrow$ & SSIM$\uparrow$ & LPIPS$\downarrow$ & PSNR$\uparrow$ & SSIM$\uparrow$ \\ 
    \midrule[1pt]

    \multicolumn{13}{c}{\textbf{G-Buffers Forward Rendering}} \\
    \midrule[0.8pt]
    \rowcolor{blue!10} DiffusionRenderer \citep{diffusionrenderer} & 0.0802 & 24.41  & 0.9172  & 0.0560 & 27.16 & 0.9354  & 0.0616 & 27.09 & 0.9288 & 0.0707 & 25.46 & 0.9126  \\ 
    \rowcolor{blue!10} Ours & \textbf{0.0488}  & \textbf{29.15}  & \textbf{0.9484}  
     & \textbf{0.0458}  & \textbf{31.59}  & \textbf{0.9586} 
     & \textbf{0.0490}  & \textbf{32.23}  & \textbf{0.9627} 
     & \textbf{0.0475}  & \textbf{30.47}  & \textbf{0.9594} \\
    \midrule[0.8pt]

    \multicolumn{13}{c}{\textbf{Random-lit Single-image Reconstruction}} \\
    \midrule[0.8pt]
    \rowcolor{green!10} RGB$\leftrightarrow$X \citep{rgbx} & 0.1605 & 15.15 & 0.8445 & 0.1349 & 15.48 & 0.8624 & 0.1199 & 16.09 & 0.8801 & 0.1271 & 14.29 & 0.8612 \\
    \rowcolor{green!10} DiLightNet \citep{zeng2024dilightnet} & 0.0949 & 21.11 & 0.8947 & 0.0650 & 23.53 & 0.9147 & 0.0507 & 25.65 & {0.9300} & 0.0523 & 24.09 & 0.9213 \\
    \rowcolor{green!10} DiffusionRenderer \citep{diffusionrenderer} & 0.0767 & {22.50} & 0.9105 & 0.0579 & 23.70 & {0.9234} & 0.0516 & 24.81 & 0.9285 & 0.0547 & 23.40 & 0.9163 \\
    \rowcolor{green!10} Ours & \textbf{0.0754} & \textbf{22.89} & \textbf{0.9116} 
     & \textbf{0.0532} & \textbf{24.68} & \textbf{0.9246} 
     & \textbf{0.0394} & \textbf{26.53} & \textbf{0.9305} 
     & \textbf{0.0368} & \textbf{25.32} & \textbf{0.9274} \\
    \midrule[0.8pt]

    \multicolumn{13}{c}{\textbf{Unknown-lit Single-image Relighting}} \\
    \midrule[0.8pt]
    \rowcolor{red!10} DiLightNet \citep{zeng2024dilightnet} & 0.1037 & 20.59 & 0.8813 & 0.0729 & 22.63 & 0.8913 & 0.0657 & 23.87 & 0.9011 & 0.0622 & 22.40 & 0.9059 \\
    \rowcolor{red!10} NeuralGrafferer \citep{jin2024neural} & 0.2675 & 14.31 & 0.7839 & 0.2548 & 14.22 & 0.7943 & 0.2108 & 14.68 & 0.8238 & 0.1767 & 15.67 & 0.8200 \\
    \rowcolor{red!10} DiffusionRenderer \citep{diffusionrenderer} & 0.0916 & 21.91 & {0.8960} & 0.0691 & 22.99 & 0.9078 & 0.0609 & 23.75 & 0.9169 & 0.0632 & 22.13 & 0.9062 \\
    \rowcolor{red!10} Ours & \textbf{0.0915} & \textbf{21.95} & \textbf{0.8972}  
     & \textbf{0.0642} & \textbf{23.47} & \textbf{0.9177}  
     & \textbf{0.0557} & \textbf{24.38} & \textbf{0.9264} 
     & \textbf{0.0465} & \textbf{22.61} & \textbf{0.9246} \\
    \midrule[0.8pt]

    \multicolumn{13}{c}{\textbf{Novel-view Relighting}} \\
    \midrule[0.8pt]
    \rowcolor{orange!10} 3DTopia-XL \citep{chen20243dtopia} & 0.1754 & 17.24 & 0.8013 & 0.1051 & 21.56 & 0.8674 & 0.0769 & 23.22 & 0.8989 & 0.0857 & 20.89 & 0.8807 \\ 
    \rowcolor{orange!10} Stable-Fast-3D \citep{sf3d} & 0.1028 & 19.43 & 0.8881 & 0.0616 & 22.07 & 0.9154 & 0.0666 & 22.26 & 0.9112 & 0.0747 & 20.17 & 0.8943 \\ 
    \rowcolor{orange!10} MeshGen \citep{chen2025meshgen} & 0.0939 & 20.15 & 0.8879 & 0.0661 & 22.87 & 0.9101 & 0.0509 & 24.15 & 0.9306 & 0.0637 & 21.43 & 0.9071 \\ 
    \rowcolor{orange!10} Hunyuan3D-2.1 \citep{zhao2025hunyuan3d} & 0.0727 & 22.30 & 0.9017 & 0.0481 & 24.89 & 0.9255 & 0.0479 & 25.47 & 0.9328 & 0.0533 & 22.26 & {0.9119} \\ 
    \rowcolor{orange!10} Ours & \textbf{0.0693} & \textbf{22.87} & \textbf{0.9023}  
     & \textbf{0.0475} & \textbf{25.53} & \textbf{0.9298}  
     & \textbf{0.0486} & \textbf{25.97} & \textbf{0.9392}  
     & \textbf{0.0502} & \textbf{22.94} & \textbf{0.9147} 
     \\
    
    \bottomrule
    \end{tabular}
    }
\vspace{-5mm}
\end{table*}

\begin{figure}[tbp]
\small
\centering
\includegraphics[width=\linewidth]{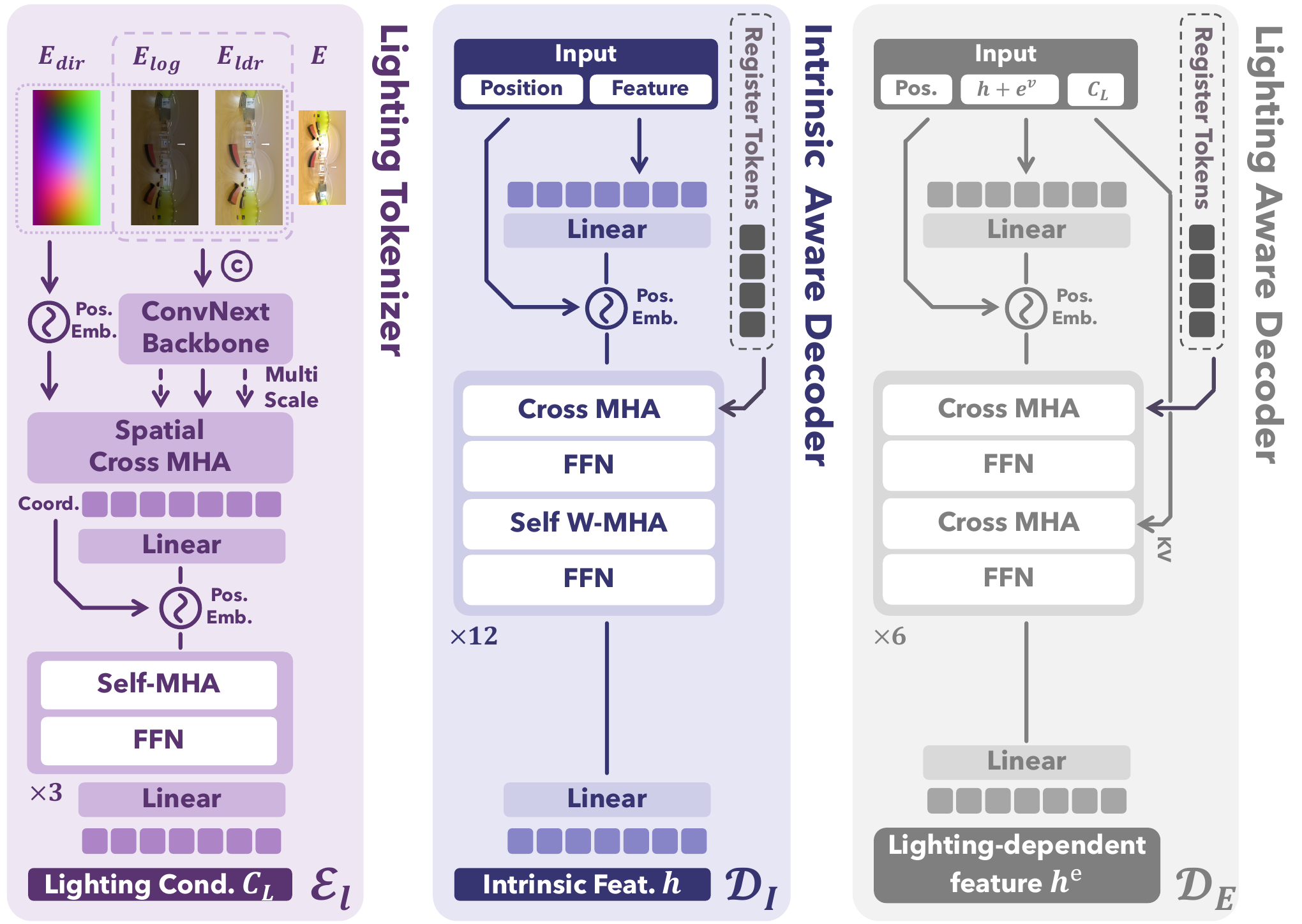}
\vspace{-15pt}
\caption{Architectures of Lighting Tokenizer, IAD, and LAD.}
\label{fig:forward}
\vspace{-5mm}
\end{figure}

\subsection{Relightable Neural 3D Gaussian Synthesis}
\label{sec:stage2}
\label{sec:rays}
The second stage synthesizes a relightable 3D Gaussian Splatting (3DGS) field $\mathcal{G}$ from LH-SLAT, conditioned on target illumination and viewpoint. Unlike optimization approaches~\cite{gao2024relightable, bi2024gs3}, we employ an efficient feed-forward decoder with two sequential modules: the \emph{Intrinsic Aware Decoder (IAD)} and the \emph{Lighting Aware Decoder (LAD)}.

\subsubsection{Intrinsic Aware Decoder (IAD)}
\label{sec:IAD}
The IAD, denoted as $\mathcal{D}_I$, aims to process LH-SLAT $Z_{\text{lh}}$ and generate a view-independent and illumination-invariant intrinsic feature $\boldsymbol{h} = \{(\boldsymbol{h}_i,\boldsymbol{p}_i)\}_{i=1}^{L}$, where $\boldsymbol{h}_i \in \mathbb{R}^{768}$. This sparse feature field $\boldsymbol{h}$ effectively decodes the underlying geometric structure and material properties of the scene. To achieve this, IAD employs a Transformer architecture akin to TRELLIS~\cite{xiang2024structured}, leveraging stacked self-shifted window attention blocks to exploit the inherent locality of structured 3D latent sequences. To further enhance the model's comprehension of global structural relationships and lighting context, a register cross-attention layer is incorporated into each block. Specifically, 16 learnable register tokens are appended to each SLAT sequence to capture global context and suppress high-frequency noise~\cite{darcet2024vision, li2025lino}. These register tokens are injected into the decoder via global cross-attention, facilitating information exchange with all latent variable tokens and enabling a coherent and globally consistent intrinsic representation $\boldsymbol{h}$.

% The IAD ($\mathcal{D}_I$) processes LH-SLAT $Z_{\text{lh}}$ to produce a view-independent and illumination-invariant intrinsic feature $\boldsymbol{h}=\{(\boldsymbol{h}_i,\boldsymbol{p}_i)\}_{i=1}^{L}$, where $\boldsymbol{h}_i\in\mathbb{R}^{768}$, encoding scene geometry and materials. It adopts a TRELLIS-like Transformer~\cite{xiang2024structured} with stacked self-shifted window attention to exploit locality in structured 3D latent sequences, and incorporates a register cross-attention layer in each block. Specifically, 16 learnable register tokens are appended to each SLAT sequence to capture global context and suppress high-frequency noise~\cite{darcet2024vision, li2025lino}. These tokens are injected into the decoder via global cross-attention, enabling coherent and globally consistent $\boldsymbol{h}$.

\subsubsection{Lighting Aware Decoder (LAD)} 
\label{sec:LAD}
The LAD, denoted as $\mathcal{D}_E$, synthesizes the final lighting-dependent features by injecting view embeddings and environmental lighting conditions, as shown in Fig. \ref{fig:pipeline}.

\textbf{Observe View Embedding}.
To explicitly model specular highlights that vary with viewing angles, we abandon the commonly used spherical harmonics and instead inject the observed view information into the learning process of LAD from the outset to enhance the model's perception of specular highlights. Along the camera ray to each voxel $\boldsymbol{p}_i$ in the world coordinate system, we record the distance $x = \{(l_i, \boldsymbol{p}_i)\}_{i=1}^{L}$, where $l_i \in \mathbb{R}$, and the ray direction $\boldsymbol{d}^w = \{({\boldsymbol{d}^w}_i, \boldsymbol{p}_i)\}_{i=1}^{L}$. We then transform $\boldsymbol{d}^w$ to the camera coordinate system using the extrinsic matrix, denoted as $\boldsymbol{d} = \{(\boldsymbol{d}_i, \boldsymbol{p}_i)\}_{i=1}^L$, where $\boldsymbol{d}_i \in \mathbb{R}^3$. We apply NeRF positional encoding and learnable positional encoding to $\boldsymbol{d}$ and $l$ voxel-wise, respectively, obtaining the view embedding:
\begin{equation*}
\boldsymbol{e}^v = \{\boldsymbol{e}^d,\boldsymbol{e}^l\} =\{(\boldsymbol{e}^{d}_i, \boldsymbol{p}_i),(\boldsymbol{e}^{l}_i, \boldsymbol{p}_i)\}_{i=1}^{L},\quad \boldsymbol{e}_i \in \mathbb{R}^{768}.
\end{equation*}
Then, we add $\boldsymbol{e}^d$ and $\boldsymbol{e}^l$ voxel-wise to $\boldsymbol{h}$ to obtain $\boldsymbol{h}^{v}$, which serves as the input to LAD.

\textbf{Lighting Tokenizer}.
We encode the high dynamic range (HDR) environment map $\mathbf{E}$ into compact lighting conditions using an HDRI encoder $\mathcal{E}_l$. 
Following~\cite{jin2024neural, diffusionrenderer, unirelight}, we decompose $\mathbf{E}$ into a tone-mapped LDR image $\mathbf{E}_{\text{ldr}}$, a normalized log-intensity map $\mathbf{E}_{\text{log}} = \log(\mathbf{E} + 1)/\mathbf{E}_{max}$, and a camera-space direction encoding $\mathbf{E}_{\text{dir}} \in \mathbb{R}^{H \times W \times 3}$.
Unlike prior works that compress the entire map via VAE, we employ a ConvNeXt backbone to extract multi-scale visual features from $\mathbf{E}_{\text{ldr}}$ and $\mathbf{E}_{\text{log}}$. 
Crucially, rather than directly compressing $\mathbf{E}_{\text{dir}}$, we first encode it via NeRF-style positional embedding~\cite{mildenhall2021nerf} and fuse it with visual features using \textbf{Spatial Cross Attention}. This mechanism acts as a learnable positional encoding, modulating visual features with explicit directional cues. 
The resulting multi-scale features are concatenated, processed with positional encoding, and passed through self-attention blocks to yield the Lighting Condition Tokens $C_L \in \mathbb{R}^{4096 \times 768}$. This design explicitly embeds directional information, facilitating editable lighting directions when switching views.

% Similar to previous works \cite{jin2024neural, diffusionrenderer, unirelight}, we obtain the low dynamic range $\mathbf{E}_{ldr}$ through Reinhard tone mapping, compute the normalized log-intensity map $\mathbf{E}_{\text{log}} = \log(\mathbf{E} + 1)/\mathbf{E}_{max}$, and generate the direction encoding $\mathbf{E}_{\text{dir}} \in \mathbb{R}^{H \times W \times 3}$ in the camera coordinate system. Differently from using VAE encoder to encode $\mathbf E$, we employ ConvNeXt to extract multi-scale visual features from the low dynamic range image $\mathbf{E}_{ldr}$ and the normalized log-intensity map $\mathbf{E}_{\text{log}}$. A key innovation is that we avoid directly compressing the direction encoding $\mathbf{E}_{\text{dir}}$. Instead, we first encode it through (NeRF) position embedding and then fuse it with visual features at multiple scales using the \textbf{Spatial Cross Attention}. The spatial cross attention acts as a learnable positional encoding, modulating the visual features at different scales via $\mathbf{E}_{\text{dir}}$ and embedding directional information into the visual representation. These multi-scale features are then concatenated along the channel dimension, further processed with positional encoding, and passed through three self-attention blocks to form the corresponding lighting condition $C_L \in \mathbb{R}^{4096 \times 768}$. This design allows us to basically edit $\mathbf{E}_{\text{dir}}$ when switching views and lighting directions.

\textbf{LAD Architecture}.
% LAD primarily consists of stacked cross-attention blocks. The lighting condition  $C_L$ is injected into the intrinsic feature $\boldsymbol{h}^{v}$ via cross-attention layers, enabling the network to be aware of the environment lighting conditions.  Similar to IAD, to enhance the perception of global illumination, we use a register cross-attention layer in each block. After LAD, we obtain the lighting-aware sparse feature $\boldsymbol{h}^e$.
LAD consists of stacked cross-attention blocks that inject lighting condition $C_L$ into intrinsic features $\boldsymbol{h}^{v}$, enabling lighting awareness. Similar to IAD, each block includes a register cross-attention layer to enhance global illumination perception. The output is the lighting-aware sparse feature $\boldsymbol{h}^e$.

% \textbf{3D Gaussian Decoding}. 
\subsubsection{Neural 3D Gaussian Splatting} 
We regress the 3DGS parameters using both the intrinsic feature $\boldsymbol{h}$ and the lighting-aware feature $\boldsymbol{h}^e$:
\begin{equation}
\label{eq:3dgs}
\begin{aligned}
    &\{(\boldsymbol{h}^v_i, \boldsymbol{p}_i)\}_{i=1}^{L} \rightarrow \{ \{(\boldsymbol{o}^k_i, \boldsymbol{b}^k_i, \gamma^k_i, \boldsymbol{m}^k_i, \boldsymbol{s}^k_i, \alpha^k_i, \boldsymbol{r}^k_i) \}_{k=1}^{K} \}_{i=1}^{L}, \\
    &\{(\boldsymbol{h}^e_i,\boldsymbol{p}_i)\}_{i=1}^{L} \rightarrow \{ \{(\boldsymbol{f}^k_i, \hat{s}^k_i, \sigma^k_i) \}_{k=1}^{K} \}_{i=1}^{L}
\end{aligned}
\end{equation}
the intrinsic feature $\boldsymbol{h}_i$ is decoded into $K$ Gaussian parameters: position offset $\boldsymbol{o}$, base color $\boldsymbol{b}$, roughness $\gamma$, metallic $\boldsymbol{m}$, scale $\boldsymbol{s}$, rotation $\boldsymbol{r}$, and opacity $\alpha$ (activated via $\tanh$ to support negative density~\cite{zhu20253d}). 
Simultaneously, the lighting-dependent feature $\boldsymbol{h}^e_i$ predicts the 48-dim color feature $\boldsymbol{f}$, lighting-specific scale $\hat{s}$, and shadow $\sigma$.
The Gaussian centers are defined as $\boldsymbol{x}^k_i = \boldsymbol{p}_i + \tanh(\boldsymbol{o}^k_i)$, with normals derived from the shortest axis of $\hat{s}$.
% We calculate the shortest axis based on the scale $\hat{s}$, and use it as the normal vector for each Gaussian primitive.
Finally, we employ a simple shallow MLP network that combines the positional encoding of the normal vector and the color feature $\boldsymbol{f}$. This network uses ReLU activation functions in its intermediate layers and an ELU activation function in its final layer to predict the radiance values for each Gaussian. Through the rasterization operation $\mathcal{M}$, we obtain the 2D HDR prediction $I^{hdr}_{target}$. 
We also render 2D base color, roughness, metallic, shadow images $I^b, I^r, I^m, I^s$.

% \begin{figure}[tbp]
% \small
% \centering
% \includegraphics[width=\linewidth]{figures/forward_rendering.pdf}
% \vspace{-6mm}
% \caption{Visual comparison of Diffusion Renderer (with G-buffer) and our LH-SLAT method for image relighting.}
% \label{fig:forward}
% \vspace{-6mm}
% \end{figure}

\textbf{Loss Function}.
\label{sec:loss}
% We supervise the training by calculating the reconstruction loss $\mathcal{L}_{hdr}$ between the rendered reference HDR image and the predicted HDR result  consists of $L_1$, LPIPS \cite{zhang2018perceptual}, and D-SSIM. 
% Following \cite{zeng2025renderformer}, to prevent minor errors in the high-light areas from dominating the $L_1$ loss, we apply a logarithmic transformation to the images. Then, we compute the LPIPS and DSSIM losses on the tonemapped versions of the two images (clamp(log I / log 2, 0, 1)). Furthermore, we calculate $L_1$ losses between the rendered reference and predicted results for material properties (including base color, roughness, and metalness) and shadows to aid in training. The total loss is a weighted sum of each individual loss, represented as: $\mathcal{L}_{stage2} = \mathcal{L}_{hdr} + \lambda_{pbr} \mathcal{L}_{pbr} + \lambda_{shadow} \mathcal{L}_{shadow}$.
% For more details refer to the appendix.
We supervise the training via an HDR reconstruction loss $\mathcal{L}_{hdr}$, which comprises $\mathcal{L}_1$, LPIPS~\cite{zhang2018perceptual}, D-SSIM and regularization terms.
Following~\cite{zeng2025renderformer}, to prevent high-intensity regions from dominating the $\mathcal{L}_1$ optimization, we apply a logarithmic transformation to the HDR images. For perceptual metrics (LPIPS and D-SSIM), we operate on tone-mapped images using $\text{clamp}(\log_2(I), 0, 1)$. 
Additionally, we impose auxiliary $\mathcal{L}_1$ supervision on material properties maps (base color, roughness, metallic), denoted as $\mathcal{L}_{pbr}$, and shadows $\mathcal{L}_{shadow}$. 
The total objective is formulated as follows: 
\begin{equation}
\small
    \mathcal{L}_{stage2} = \mathcal{L}_{hdr} + \lambda_{pbr} \mathcal{L}_{pbr} + \lambda_{shadow} \mathcal{L}_{shadow}.
    \label{eq:loss}
\end{equation}

\section{Experiments}
\subsection{Implementation Details}
% We refer to implementation details in the Appendix.
\label{sec:implementation}

% More training details refer to the supplyment.
Please refer to the Supplementary Material for comprehensive implementation details.  
% \textbf{Training details}. The training pipeline is executed on 4 NVIDIA H100 80GB HBM3 GPUs. The initial stage involves training the flow model, where we employ LoRA initialized using the PEFT \cite{peft}. The LoRA configuration consists of a rank of 512 and a scaling factor of 512. LoRA is applied to query, key-value, output projection, and combined query-key-value modules within the attention mechanism. The AdamW optimizer~\cite{loshchilov2018decoupled} is used with a learning rate of $1.0 \times 10^{-4}$. The first stage requires approximately one day for completion. In the second stage, we utilize the AdamW~\cite{loshchilov2018decoupled} optimizer with a batch size of 48 and a linear warmup learning rate of $1.0 \times 10^{-4}$ over 5,000 steps, followed by a cosine decay schedule. An end-to-end joint training of the IAD, LAD, and $\mathcal E_\text{l}$ is performed. Training acceleration is achieved through the implementation of Flash-Attention 3 \cite{flash3} and the gsplat \cite{ye2025gsplat}. Initially, the model is trained with all loss components for 400K iterations, requiring approximately 8 days. Subsequently, the PBR rendering loss is removed, and training continues for an additional 100K iterations, taking approximately 2 days.

% \begin{figure}[tbp]
% \small
% \centering
% \includegraphics[width=0.95 \linewidth]{figures/exp.pdf}
% \vspace{-3mm}
% \caption{Schematic illustration of four distinct sub-tasks.}
% \label{fig:subtask}
% \vspace{-6mm}
% \end{figure}

\begin{figure}[tbp]
\small
\centering
\includegraphics[width=\linewidth]{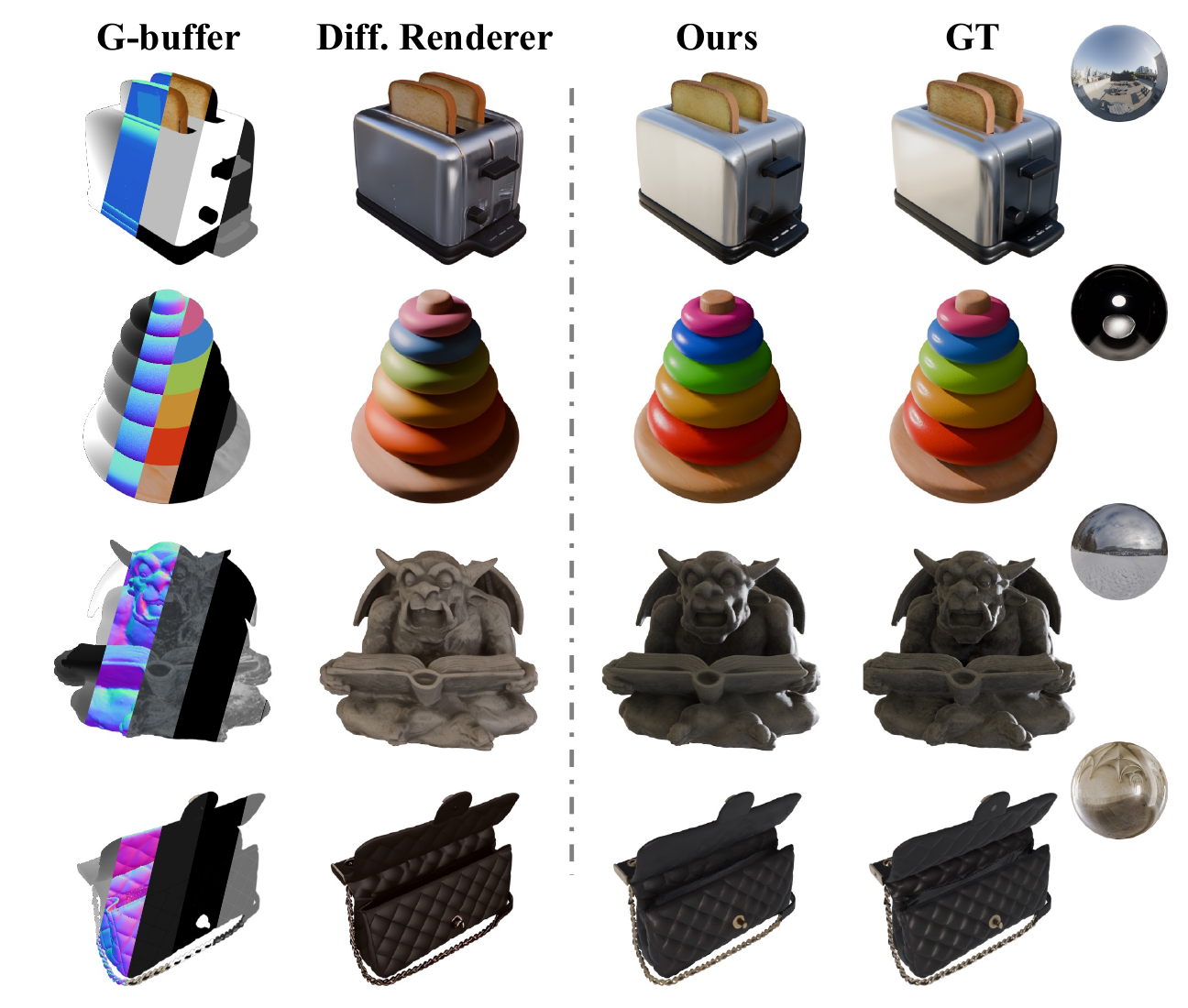}
\vspace{-6mm}
\caption{Visual comparison of Diffusion Renderer (with G-buffer) and our LH-SLAT method for image relighting.}
\label{fig:forward}
\vspace{-6mm}
\end{figure}

\textbf{Training Data}. 
Our training dataset comprises 87K 3D assets with physically-based rendering (PBR) textures, curated from the Objaverse-XL dataset. These assets are illuminated using 2K High Dynamic Range Images (HDRIs), each at 4K resolution, used as environment maps. We normalized the assets to fit within a bounding box of $[-0.5, 0.5]$. The first training stage involves rendering 150 viewpoints under normalized lighting to extract illumination-invariant structural latent representations. For input images under unknown illumination, camera poses are sampled with yaw within ±45 degrees and pitch from -10 to 45 degrees, oriented towards the object's center, and with field of view (FOV) and radius following \cite{xiang2024structured}. Unknown illumination is modeled with (1) six area lights uniformly distributed on a sphere, (2) 1-3 area lights randomly sampled within the camera's hemisphere, or (3) a random, Z-axis-rotated environment map. Area light intensities are sampled uniformly between 300 and 700 (units), distances between 5 and 8 units. 
In the second stage, we re-light objects using randomly rotated environment maps as supervision, with a fixed FOV of $40^\circ$. We randomly and uniformly sample 12 camera viewpoints on a sphere of radius 2.0, where each viewpoint is rendered under 16 different illumination conditions. All data generation across both stages utilizes the Blender EEVEE Next engine~\cite{eevee} with raytracing enabled.

\textbf{Task Definitions And Baselines}. We evaluate our method on two fundamental tasks: single-view forward rendering and novel view relighting from single-image to Relightable 3D. 
We evaluate the consistency between the rendered outputs and the ground truth reference images. 
The former involves single-view forward rendering with input G-buffers (such as normals, material, and depth information), image reconstruction from a single-image under random lighting, and relighting of a single image under unknown lighting. For single-view forward rendering, we compare against recent state-of-the-art neural rendering methods RGB$\leftrightarrow$X \cite{rgbx}, neural-gaffer \cite{jin2024neural}, DiLightNet \cite{zeng2024dilightnet}, and Diffusion-render \cite{diffusionrenderer}. For novel view relighting, we compare against recent open-source methods that support single-image to 3D generation with PBR materials, including Huyuan3D-2.1~\cite{zhao2025hunyuan3d} (HY3D 2.1), MeshGen~\cite{chen2025meshgen}, 3DTopia-XL~\cite{chen20243dtopia}, and SF3D~\cite{sf3d}. The schematic diagram for the four subtasks is illustrated in Fig. \ref{fig:subtask}. We additionally present qualitative results for PBR material estimation in comparison with HY3D 2.1.

\textbf{Evaluation Metric}. We use PSNR, SSIM \cite{wang2004image} and LPIPS \cite{zhang2018perceptual} to measure the quality of the rendering.

\begin{figure}[tbp]
\small
\centering
\includegraphics[width=\linewidth]{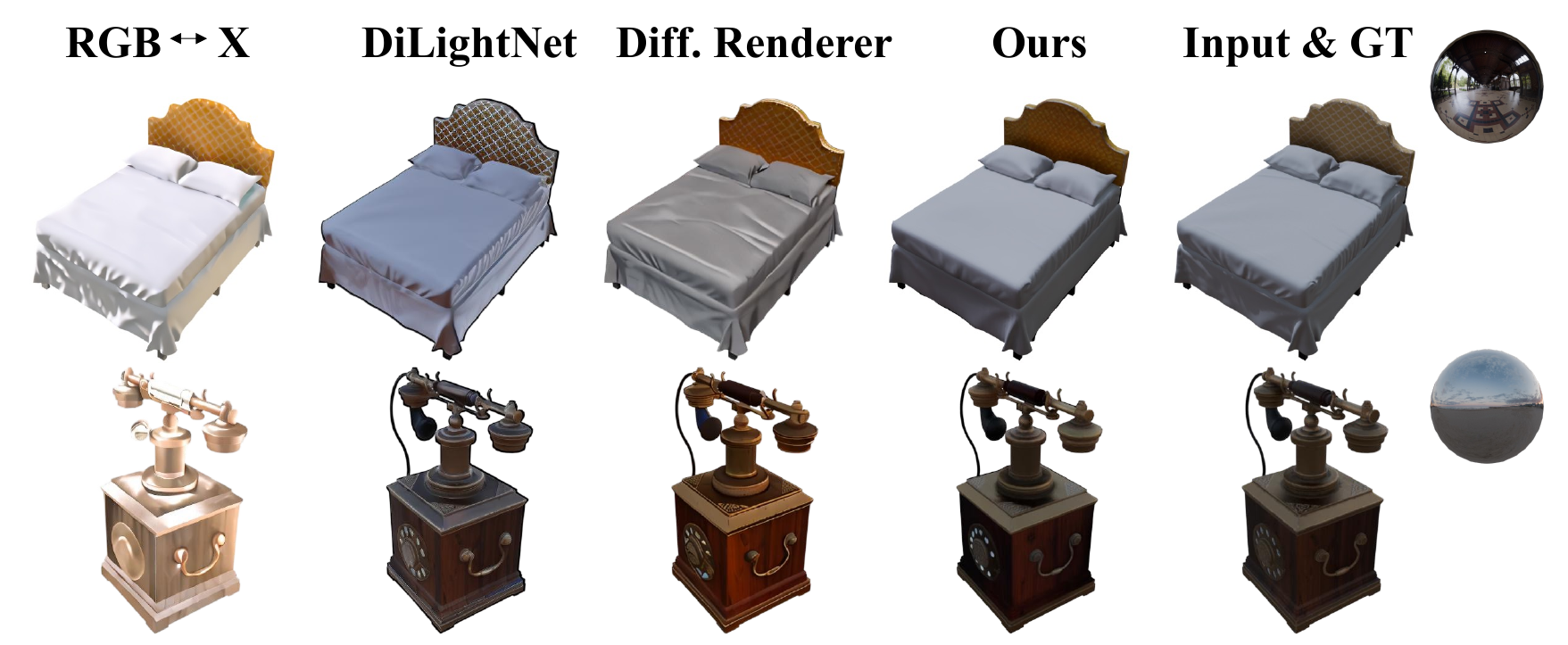}
  % \captionsetup{font=\smaller}
\caption{Visual comparison of image reconstruction.}
\vspace{-2mm}
\label{fig:recon}
\end{figure}

\begin{figure}[tbp]
\small
\centering
\includegraphics[width=\linewidth]{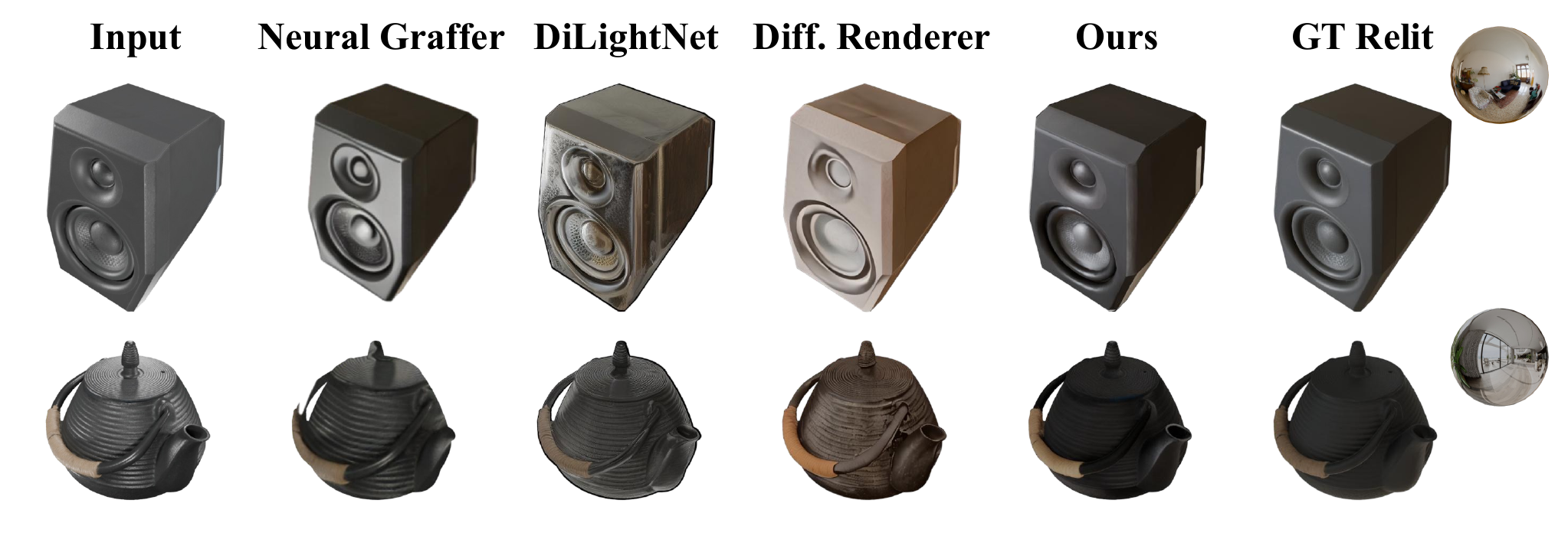}
% \captionsetup{font=\smaller}
\vspace{-15pt}
\caption{Visual comparison of relighting.}
\label{fig:relit}
\vspace{-3mm}
\end{figure}

\textbf{Evaluation Datasets}. 
% We randomly select 800 objects from the training data to create a test set, ensuring that these objects were not seen by the model during training.
We construct a test set by randomly selecting 800 unseen objects from our training data.
% To validate the generalizability of our method, we utilize the Aria Digital Twin (ADT)~\cite{pan2023aria} and Digital Twin Catalog (DTC)~\cite{dong2025digital} datasets as out-of-domain datasets. These datasets provide comprehensive resources for 3D object modeling, featuring a vast library of highly detailed, photorealistic models with sub-millimeter accuracy. We further incorporate the Glossy Synthetic dataset~\cite{nero}, which provides 3D assets, and expand it with additional assets sourced from the BlenderKit \footnote{https://www.blenderkit.com/}. We also modify rendering nodes to utilize the Principled BSDF shader \footnote{https://www.blender.org/}.
% We construct a test set by randomly selecting 800 unseen objects from our training data. 
To validate generalization capability, we evaluate on out-of-domain datasets: Aria Digital Twin (ADT)~\cite{pan2023aria} and Digital Twin Catalog (DTC)~\cite{dong2025digital}, which feature high-fidelity photorealistic models with sub-millimeter accuracy. We also incorporate the Glossy Synthetic dataset~\cite{nero} and additional assets from BlenderKit\footnote{https://www.blenderkit.com/}, modifying rendering nodes to utilize the Principled BSDF shader\footnote{https://www.blender.org/}.

\subsection{Single-view Forward Rendering}

% \begin{figure}
%     \centering
%     \includegraphics[width=0.5\linewidth]{iclr2026/figures/forward_rendering.pdf}
%     \small\caption{Visual comparison of Diffusion Renderer with Gbuffer/LH-SLAT for image relighting.}
%     \label{fig:forward}
% \end{figure}

% \begin{figure}
%     \centering
%     \includegraphics[width=0.5\linewidth]{iclr2026/figures/recon.pdf}
%     \small\caption{Visual comparison of image reconstruction accuracy under corresponding lighting.}
%     \label{fig:recon}
% \end{figure}

% \begin{figure}
%     \centering
%     \includegraphics[width=0.5\linewidth]{iclr2026/figures/relighting.pdf}
%     \caption{Qualitative comparison: our method better reconstructs color and specular reflections than baselines.}
%     \label{fig:relit}
% \end{figure}

% \begin{figure}[tbp]
% \small
% \centering
% \begin{minipage}{0.48\textwidth}
% \centering
% \includegraphics[width=\linewidth]{figures/forward_rendering.pdf}
%   % \captionsetup{font=\smaller}
% \caption{Visual comparison of Diffusion Renderer with Gbuffer/LH-SLAT for image relighting.}
% \label{fig:forward}
% \end{minipage}\hfill
% \begin{minipage}{0.48\textwidth}
% \centering
% \includegraphics[width=\linewidth]{figures/recon.pdf}
%   % \captionsetup{font=\smaller}
% \caption{Visual comparison of image reconstruction.}
% \label{fig:recon}

% \includegraphics[width=\linewidth]{figures/relighting.pdf}
% % \captionsetup{font=\smaller}
% \caption{Visual comparison of relighting.}
% \label{fig:relit}
% \end{minipage}

% \end{figure}

\textbf{G-buffers Forward Rendering}.
As shown in Fig.~\ref{fig:forward}, we compare against Diffusion Renderer using ground truth G-buffers and LH-Slat 
% (with Base Color SLAT)
, bypassing the single-image-to-intermediate representation step. Our method demonstrates superior accuracy in shadow and highlight distribution (e.g., the toy's specular highlight and the sculpture's shadow detail), likely due to our explicit 3D structural information. Furthermore, we accurately capture material reflections of ambient light, as illustrated by the stainless steel. Quantitatively, our method significantly outperforms baselines across four datasets in Tab.~\ref{tab:comp}.

% \begin{figure}[tbp]
% \small
% \centering
% \includegraphics[width=\linewidth]{figures/pbr.pdf}
% % \captionsetup{font=\smaller}
% \vspace{-15pt}
% \caption{Visual comparison of PBR material estimation.}
% \label{fig:relit}
% \vspace{-3mm}
% \end{figure}

% As shown in Fig.~\ref{fig:forward}, we compare our method against Diffusion Renderer (using ground truth G-buffers) and our own variant using Base Color SLAT. Our approach demonstrates superior accuracy in shadow casting and highlight distribution, exemplified by the specular highlights on the toy and the intricate shadow details on the sculpture. This suggests that our explicit 3D structural priors are more effective than 2D feature maps. Furthermore, our method accurately captures material reflections of ambient light (e.g., the stainless steel object). Quantitatively, our method significantly outperforms baselines across four datasets, as reported in Fig.~\ref{fig:recon} (Quantitative table reference needed here if distinct from Fig).

\begin{table}[tbp]
\centering
% \caption{Ablation Studies}
\label{tab:combined_ablation}
\small
  \centering
  \vspace{-3mm}
  \caption{Ablation study on the number of blocks for $\mathcal{D}_I$ and $\mathcal{D}_E$.}
  \label{tab:model_depths}
  \vspace{-2mm}
  \begin{tabular}{lccccc}
    \toprule
    Num & PSNR & SSIM & LPIPS & $\mathcal{D}_E$ Param. & FPS \\
    \midrule
    12 + 1  & 31.56 & 0.9608 & 0.0508 & 12.65M & 48\\
    12 + 3  & 32.35 & 0.9635 & 0.0474 & 31.55M & 38\\
    \rowcolor{blue!10} 12 + 6  & 32.54 & 0.9649 & 0.0442 & 59.8M & 30 \\
    12 + 9  & 32.56 & 0.9645 & 0.0439 & 88.23M & 23 \\
    0 + 18  & 29.43 & 0.9245 & 0.0624 & 173.25M & 10 \\
    \bottomrule
  \end{tabular}
  \vspace{-5pt}
\end{table}

\begin{table}[tbp]
\small
  \centering
  \caption{Ablation study on decoder input SLAT types.}
  \vspace{-2mm}
  \label{tab:input}
  \begin{tabular}{lccc}
    \toprule
    SLAT types & PSNR & SSIM & LPIPS \\
    \midrule
    shaded & 28.95 &  0.9281 &  0.0813 \\
    base color & 30.38 & 0.9541  & 0.0564 \\
    \rowcolor{blue!10} LH & 32.02 & 0.9631 & 0.0494 \\
    \rowcolor{blue!10} LH + base color & 32.54 & 0.9649 & 0.0442 \\
    \bottomrule
  \end{tabular}
  \vspace{-4mm}
\end{table}

\begin{table}[tbp]
\centering
% \caption{Ablation Study on Architectures}
\label{tab:combined_arch}
\begin{minipage}{0.48\textwidth}
  \centering
  \includegraphics[width=0.9\linewidth]{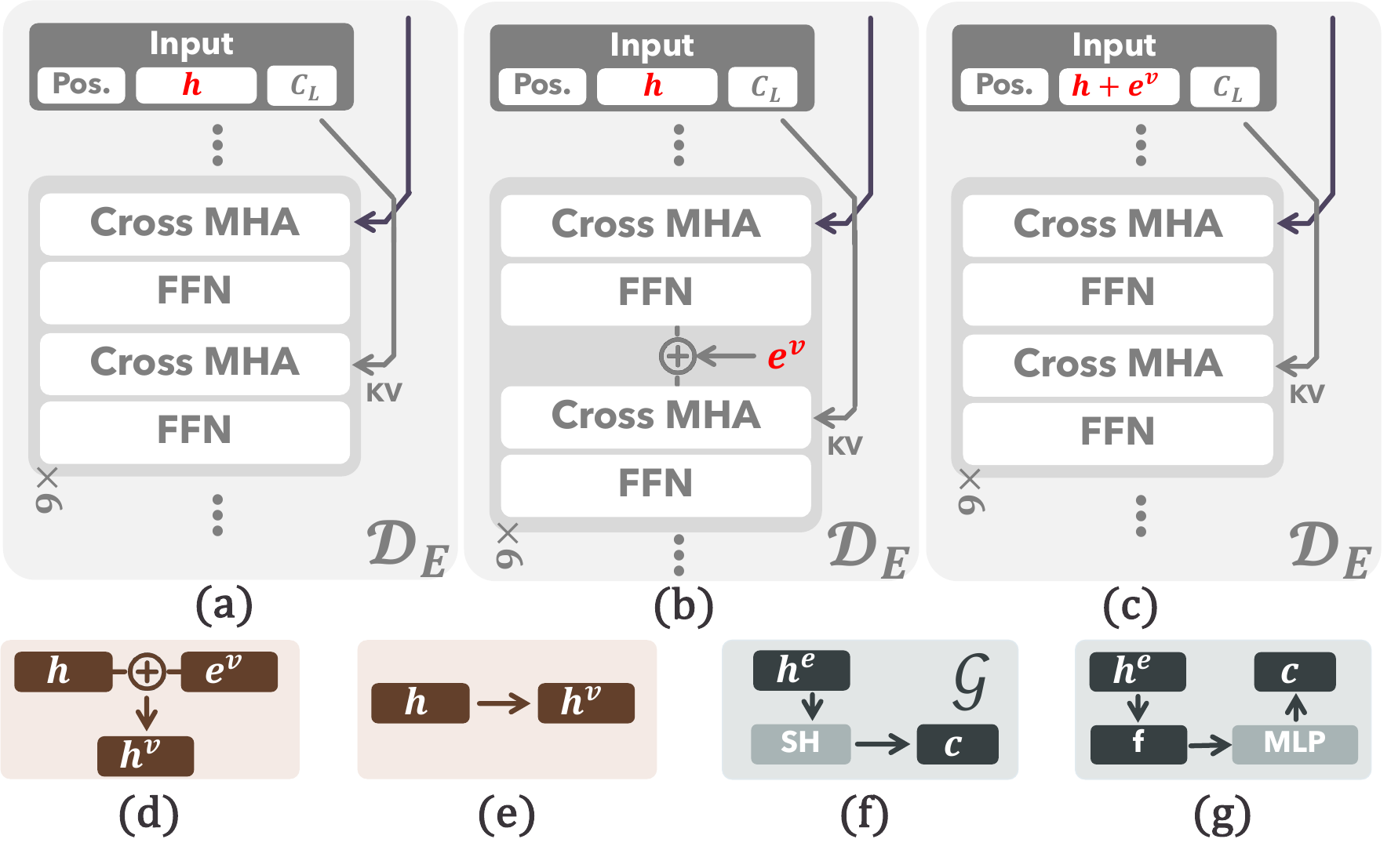}
  \vspace{-3mm}
  \captionof{figure}{Different designs for the feedforward network $\mathcal{D}$.}
  \label{fig:ab_arch}
\end{minipage}\hfill
\vspace{2pt}
\begin{minipage}{0.48\textwidth}
 \small
  \centering
  \captionof{table}{Performance Comparison of Different Architectures.}
  \label{tab:architectures}
  \vspace{-2mm}
  \begin{tabular}{lccc}
    \toprule
    Arch & PSNR & SSIM & LPIPS \\
    \midrule
    a + e + f & 29.82 &  0.9472 & 0.0642 \\
    a + e + g & 30.66 & 0.9524 & 0.0515 \\
    a + d + g  & 31.96 & 0.9597 & 0.0492 \\
    b + d + g & 32.43 & 0.9628 & 0.0472 \\
    \rowcolor{blue!10} c + d + g (ours) &32.54 & 0.9649 & 0.0442 \\
    \bottomrule
  \end{tabular}
\end{minipage}
\vspace{-2mm}
\end{table}

\textbf{Random-lit Single-image Reconstruction}. 
% As shown in Fig.~\ref{fig:recon}, our method provides improved image reconstruction compared to the baseline and outperforms the Diffusion Renderer's estimated intrinsic property approach. Quantitative evaluations in Tab.~\ref{tab:comp} demonstrate our method's advantage across the majority of metrics.
% As shown in Figs.~\ref{fig:recon},~\ref{fig:supprec1} and~\ref{fig:supprec2}, our method achieves higher reconstruction fidelity compared to baselines. Specifically, Diffusion Renderer and RGB-X misestimate materials, while DiLightNet exhibits color shifts. Quantitative evaluations in Tab.~\ref{tab:comp} confirm our method's advantage across all metrics.
As shown in Figs.\ref{fig:recon},\ref{fig:supprec1} and~\ref{fig:supprec2}, our method achieves higher reconstruction fidelity than the baselines, producing more visually consistent and geometrically faithful results. Specifically, Diffusion Renderer and RGB-X misestimate materials, while DiLightNet exhibits color shifts. The quantitative evaluations in Tab.~\ref{tab:comp} further confirm the superiority of our approach: we achieve better performance across all metrics, demonstrating improved accuracy and robustness under varying conditions.

\textbf{Unknown-lit Single-image Relighting}. 
Our method achieves more accurate highlights and color in relit images with unknown lighting, compared to other methods, as shown in Fig.~\ref{fig:relit},~\ref{fig:supprelit1} and~\ref{fig:supprelit2}. For example, observe the highlights on the speaker cones (first row) and the teapot color (second row). Tab.~\ref{tab:comp} quantitatively demonstrates the superiority of our method.
% As shown in Fig.\ref{fig:relit}, our method produces relit images with superior highlight accuracy and color fidelity under unknown lighting conditions compared to competing methods. This is evident in the highlights on the speaker cones (first row) and the preservation of the teapot's original color (second row). Furthermore, quantitative results in Tab.\ref{tab:comp} demonstrate the robust generalization capabilities of our approach.

\textbf{Novel-view Relighting}. We benchmark our full pipeline (single-image to relightable 3D) against state-of-the-art generation methods. While other methods typically reconstruct a mesh and rely on Blender for relighting, we directly generate a relightable 3D Gaussian field. 
% As shown in Fig.~\ref{fig:teaser} and \ref{fig:suppview}, compared to textured mesh methods~\cite{zhao2025hunyuan3d, chen20243dtopia, chen2025meshgen, sf3d} generated from images, our approach achieves more realistic lighting–material interactions. 
As shown in Figs.~\ref{fig:teaser} and \ref{fig:suppview}, our method achieves more realistic lighting–material interactions than image-based textured mesh methods~\cite{zhao2025hunyuan3d, chen20243dtopia, chen2025meshgen, sf3d}.
Quantitative results in Tab.~\ref{tab:comp} demonstrate improvements over existing 3D generation baselines. 
% We compared our method for novel view relighting and reconstruction against state-of-the-art image-to-3D methods supporting PBR materials. Given a single image, we generate a relightable 3D Gaussian field and use neural rendering for relighting, while other methods reconstruct the 3D model and use Blender. Fig.~\ref{fig:teaser} shows that, with the same mesh, our method achieves more accurate lighting and material interactions than Hunyuan3D. Our quantitative results in Tab.~\ref{tab:comp} demonstrate significant improvements over other 3D generation methods.

% \begin{table}[tbp]
% \centering
% % \caption{Ablation Study on Architectures}
% \label{tab:combined_arch}
% \begin{minipage}{0.48\textwidth}
%   \centering
%   \includegraphics[width=0.9\linewidth]{figures/arch.pdf}
%   \vspace{-4mm}
%   \captionof{figure}{Different designs for the feedforward network $\mathcal{D}$.}
%   \label{fig:ab_arch}
% \end{minipage}\hfill
% \vspace{2pt}
% \begin{minipage}{0.48\textwidth}
%  \small
%   \centering
%   \captionof{table}{Performance Comparison of Different Architectures.}
%   \label{tab:architectures}
%   \vspace{-2mm}
%   \begin{tabular}{lccc}
%     \toprule
%     Arch & PSNR & SSIM & LPIPS \\
%     \midrule
%     a + e + f & 29.82 &  0.9472 & 0.0642 \\
%     a + e + g & 30.66 & 0.9524 & 0.0515 \\
%     a + d + g  & 31.96 & 0.9597 & 0.0492 \\
%     b + d + g & 32.43 & 0.9628 & 0.0472 \\
%     \rowcolor{blue!10} c + d + g (ours) &32.54 & 0.9649 & 0.0442 \\
%     \bottomrule
%   \end{tabular}
% \end{minipage}
% \vspace{-4mm}
% \end{table}

\begin{figure}[tbp]
\small
\centering
\includegraphics[width=\linewidth]{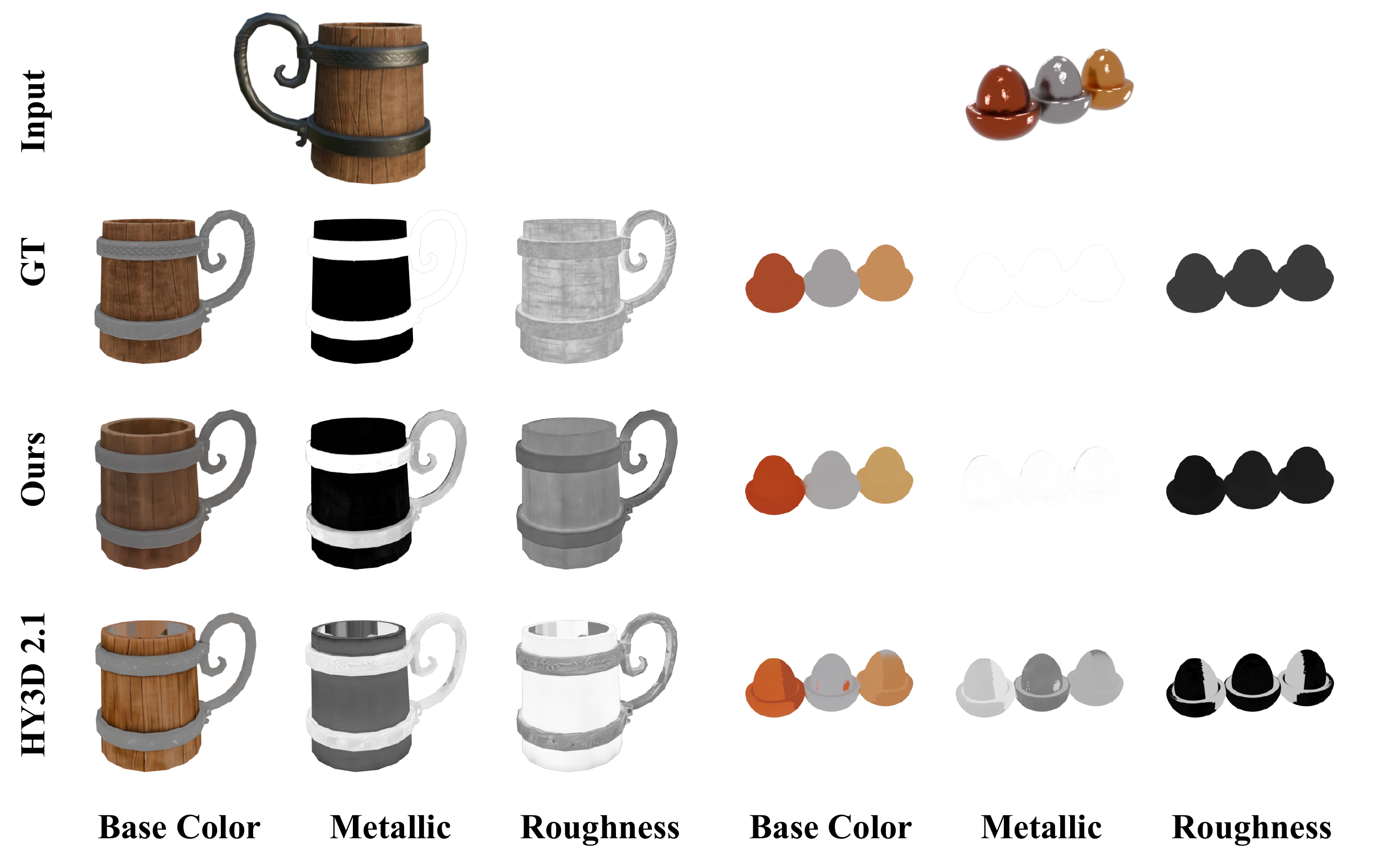}
% \captionsetup{font=\smaller}
\vspace{-2pt}
\caption{Visual comparison of PBR material estimation with HY3D2.1 and our method.}
\label{fig:pbr}
\vspace{-3mm}
\end{figure}

\begin{figure*}[t!]
  \centering
  % \fbox{\rule{0pt}{0.5in} \rule{0.9\linewidth}{0pt}}
  % \vspace{-8pt}
  \includegraphics[width=0.95\linewidth]{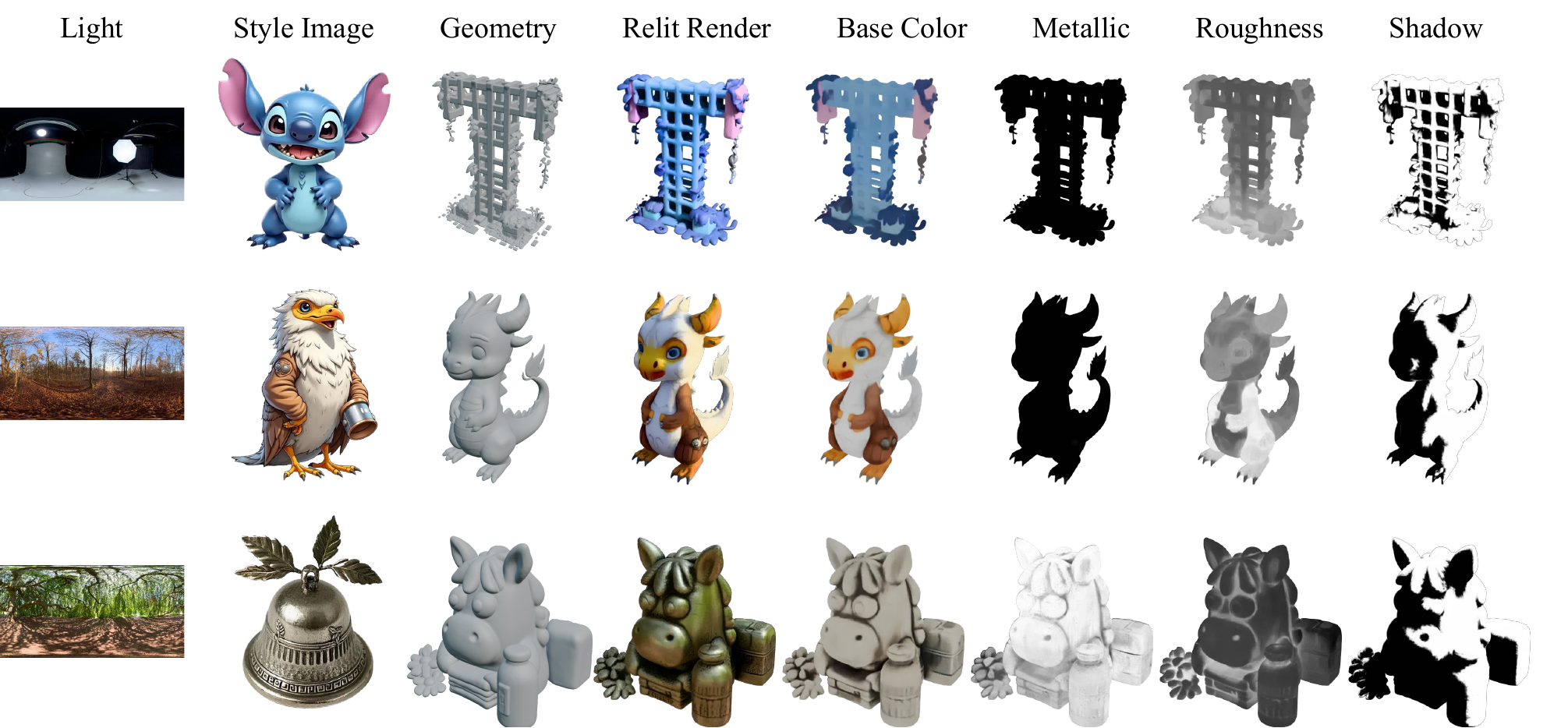}
  % \vspace{-10pt}
   \caption{\textbf{Texture Style Transfer.} Given the geometry and a target-style image as guidance, our method can generate semantically consistent stylized textures and support photorealistic neural relighting.}
   \label{fig:style}
   % \vspace{-12pt}
\end{figure*}

\begin{figure}[t]
  \centering
  % \fbox{\rule{0pt}{0.5in} \rule{0.9\linewidth}{0pt}}
  % \vspace{-8pt}
  \includegraphics[width=0.9\linewidth]{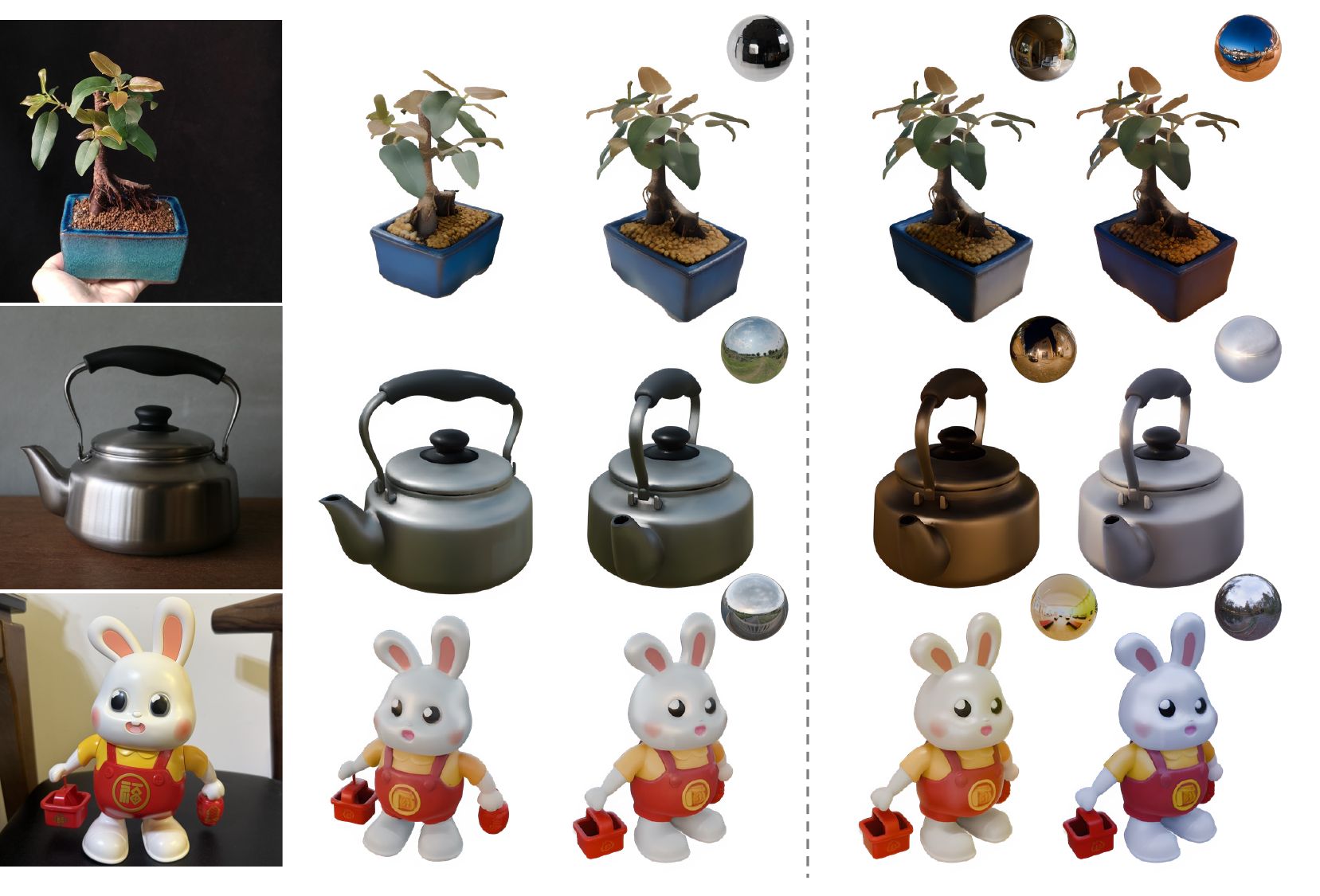}
  % \vspace{-10pt}
   \caption{Relighting results from real-world single images.}
   \label{fig:onecol}
   \vspace{-6pt}
\end{figure}

\textbf{PBR Materials Estimation}. 
Fig.~\ref{fig:pbr} demonstrates that our method surpasses the open-source SOTA model, HY3D 2.1, in material recovery. Hunyuan3D relies on multi-view diffusion, which often introduces view-inconsistent artifacts (e.g., blurred edges on the wooden cup). In contrast, our LH-SLAT preserves 3D consistency and retains crucial light-material interaction cues. For instance, HY3D 2.1 misclassifies wood as metal, resulting in erroneous metallic artifacts on the eggs, whereas our method correctly recovers the material properties.

\subsection{Applications}

\textbf{Texture Style Transfer.}
As shown in Fig.~\ref{fig:style}, given the target-style image in the second column and an arbitrary geometry, we convert the geometry as coordinates, while deriving the LH-SLAT from the stylized image using the flow models $f_{s}$ and $f_{\theta}$. This enables semantically consistent style transfer and material estimation. For instance, the dragon’s mouth in the second row corresponds to the beak style in the reference image, and the third-row horse metalness map matches the metallic appearance implied by the target style. Moreover, benefiting from LH-SLAT, we can produce photorealistic neural relighting in the fourth column under a given illumination condition, for example, the green reflections from the tree and metallic highlights in the third row.

\textbf{Real-World Images.} As shown in Fig.~\ref{fig:onecol}, to demonstrate generalization to real-world data, our method successfully removes baked-in lighting from both internet images (rows 1–2) and real mobile photos (row 3), enabling effective relighting under novel views. It further exhibits strong robustness in handling specular highlights and capturing lighting directionality (rows 2–3).

% Fig.~\ref{fig:pbr} illustrates that our method surpasses the open-source state-of-the-art model, HY3D 2.1, in material recovery accuracy. Hy3D utilizes multi-view diffusion that can introduce viewpoint inconsistencies, evident at the edges of the wooden cup. Our approach uses SLAT to better preserve 3D consistency. LH-SLAT retains interaction information between materials and uniform lighting, enhancing texture recovery. In contrast, HY3D 2.1 misclassifies wood as metal, leading to inaccuracies in estimating the metallic properties of eggs.

% \begin{figure}
%     \centering
%     \includegraphics[width=0.5\linewidth]{iclr2026/figures/arch.pdf}
%     \caption{Different component designs for the feedforward network $\mathcal{D}$.}
%     \label{fig:ab_arch}
% \end{figure}

% \begin{table}[h!]
% \centering
% \caption{}
% \begin{tabular}{lccc}
% \toprule
% Arch & PSNR & SSIM & LPIPS \\
% \midrule
% a + e + f & 29.82 &  0.9472 & 0.0642 \\
% a + e + g & 30.66 & 0.9524 & 0.0515 \\
% a + d + g  & 31.96 & 0.9597 & 0.0492 \\
% b + d + g & 32.43 & 0.9628 & 0.0472 \\
% c + d + g &32.54 & 0.9649 & 0.0442 \\
% \bottomrule
% \end{tabular}
% \label{tab:architectures}
% \end{table}

\subsection{Ablation Study.}
\noindent We perform ablation studies on our test set, investigating the Variants of $\mathcal{D}$ and input SLAT types. 

\textbf{Variants of $\mathcal{D}$}.
Tab.~\ref{tab:model_depths} indicates that increasing the depth of $\mathcal{D}_E$ improves quality but reduces inference speed; we therefore select 6 layers to strike a balance between efficiency and performance. Relying solely on the LAD $\mathcal{D}_E$ leads to a significant decline in relighting performance, consistent with~\citep{zeng2025renderformer}. Furthermore, Fig.~\ref{fig:ab_arch} demonstrates that injecting camera view information to identify which lighting tokens should be attended to, \textit{prior} to lighting baking, significantly enhances relighting results compared to baking global lighting first. This design allows for more effective capture of geometric and lighting variations, boosting the performance of $\mathcal{D}_E$ (Tab.~\ref{tab:architectures}).

% LH-SLAT, containing richer and more consistent lighting interaction information, performs better than base color SLAT and shaded SLAT $Z_s$. The $Z_s$ complicates the relighting task due to unknown lighting. However, base color SLAT complements LH-SLAT, further improving performance when used together (Tab.~\ref{tab:input}). 
\textbf{Input Types}.
We analyze the effect of different input latent representations on the decoder $\mathcal D$ in Tab.~\ref{tab:input}. LH-SLAT, which encodes rich and consistent lighting interaction information, outperforms both Base Color SLAT and Shaded SLAT ($Z_s$). The use of $Z_s$ complicates relighting due to the entanglement of unknown lighting. However, Base Color SLAT serves as a valuable complement to LH-SLAT; their combination yields the best performance.

% \textbf{Network architecture}.

\section{Conclusion}
We propose a compact multi-stage framework for relightable 3D generation, enabling consistent high-fidelity reconstruction and realistic relighting. Experiments show improved quantitative and perceptual results over strong baselines, and ablations confirm each component's contribution. Although evaluated on controlled captures with moderate compute, the approach suggests clear directions for in-the-wild and dynamic scenes and for efficiency and generalization improvements. We hope this work advances practical neural relighting and reconstruction.

\textbf{Acknowledgments}: 
This research was supported by the Beijing Natural Science Foundation (L244043), the Zhejiang Provincial Natural Science Foundation (LD24F020007).

{
    \small
    \bibliographystyle{ieeenat_fullname}
    \bibliography{main}
}

\clearpage
\appendix
\setcounter{page}{1}

\maketitlesupplementary

\section{More Implementation Details} 
\label{sec:implement}
\subsection{Different Single-Image Relighting Paradigms}
To further elucidate the architectural advantages of NeAR, we present a structured comparison of three representative paradigms in Fig.~\ref{fig:paras}.

\paravspace

\paragraph{2D-based Intrinsic Decomposition \cite{rgbx, diffusionrenderer}):} As shown in Fig.~\ref{fig:paras}(a-b), these methods decompose a single image into 2D PBR maps (e.g., albedo, roughness, normals). However, lacking an underlying 3D representation, they struggle to disentangle view-dependent specular highlights from base color, and fail to support novel-view rendering and accurate shadow modeling.

\paravspace

\paragraph{Single Image to 3D Textured Mesh ~\cite{zhao2025hunyuan3d, chen2025meshgen, chen20243dtopia}):} Typical 3D generative models (Fig.~\ref{fig:paras}(c)) follow a decoupled paradigm, where asset construction (geometry and PBR textures) is fully separated from rendering. While compatible with standard graphics pipelines (e.g., Blender), they rely on highly ill-posed PBR inversion, often leading to material ambiguity (e.g., misclassifying metallic surfaces as diffuse).

\paravspace

\paragraph{Our Coupled NeAR Stack:} In contrast, NeAR adopts a “homogenize-then-synthesize” paradigm. By lifting single-image inputs under arbitrary lighting into a Lighting-Homogenized SLAT (LH-SLAT), we construct an illumination-invariant neural asset that serves as a stable rendering substrate, preserving cues of geometry, uniform lighting, and material interactions. On top of this, a coupled neural renderer learns to interpret these homogenized latents, enabling the synthesis of complex light–material interactions.

\subsection{Implementation Details}
\paragraph{Training Details.} 

We conduct all training experiments on four NVIDIA H100 80GB HBM3 GPUs.

In the \textbf{LH-SLAT Reconstruction \& Generation phase} (\S\ref{sec:stage1}), we fine-tune a rectified flow model equipped with LoRA~\citep{hu2022lora} to normalize shaded SLATs from arbitrary images into LH-SLATs. The goal of this phase is to learn a mapping, $f_\theta$, that transforms light-dependent shaded SLAT representations into light-homogenized counterparts. To achieve this efficiently while preserving the prior knowledge of the original flow model $f_s$~\citep{xiang2024structured}, we initialize LoRA using PEFT~\cite{peft}. We configure LoRA with a rank of 512 and a scaling factor, further integrating rslora~\cite{rslora} to enhance training stability. LoRA adaptors are applied to the query, key, value, and output projection modules within the attention mechanism. We optimize the model using AdamW~\cite{loshchilov2018decoupled} with a learning rate of $1 \times 10^{-4}$. This training phase takes approximately two days to complete.

In the \textbf{Relightable Neural 3DGS Synthesis phase} (\S\ref{sec:stage2}), we employ the AdamW optimizer with a batch size of 48. The learning rate is warmed up linearly to $1 \times 10^{-4}$ over the first 5K steps, followed by a cosine decay schedule. We perform end-to-end joint training on the IAD, LAD, and the Lighting-Aware tokenizer (denoted as $E_l$). To accelerate training, we leverage Flash-Attention 3~\cite{flash3} and gsplat~\cite{ye2025gsplat}. The model is trained for 500K iterations across all loss components, requiring approximately 10 days. Additionally, we investigated the incorporation of geometric constraint losses, specifically normal and depth losses. However, we observed that adding these regularizations degraded both convergence speed and rendering quality, suggesting a trade-off between geometric constraints and rendering fidelity within the 3DGS framework~\cite{pgsr, gof}.

% The initial stage involves training the flow model, where we employ LoRA initialized using the PEFT \cite{peft}. The LoRA configuration consists of a rank of 512 and a scaling factor of 512. LoRA is applied to query, key-value, output projection, and combined query-key-value modules within the attention mechanism. The AdamW optimizer~\cite{loshchilov2018decoupled} is used with a learning rate of $1.0 \times 10^{-4}$. The first stage requires approximately one day for completion. In the second stage, we utilize the AdamW~\cite{loshchilov2018decoupled} optimizer with a batch size of 48 and a linear warmup learning rate of $1.0 \times 10^{-4}$ over 5,000 steps, followed by a cosine decay schedule. An end-to-end joint training of the IAD, LAD, and $\mathcal E_\text{l}$ is performed. Training acceleration is achieved through the implementation of Flash-Attention 3 \cite{flash3} and the gsplat \cite{ye2025gsplat}. Initially, the model is trained with all loss components for 400K iterations, requiring approximately 8 days. Subsequently, the PBR rendering loss is removed, and training continues for an additional 100K iterations, taking approximately 2 days.

\paravspace

\paragraph{Inference Details.} 
% Given a single image with unknown lighting $I_{in}$ and a high dynamic range environment map $E$, and since our method is independent of geometric generation, we first reconstruct a geometric mesh $m$ from the image using the default settings of huuyuan3D 2.1 (Hy3D 2.1) \cite{zhao2025hunyuan3d}, which is then voxelized to serve as the coordinates for the structurally sparse voxel feature SLAT. Following the approach of trellis3D, we use a pre-trained SLAT flow model $f_s$ \citep{xiang2024structured} to generate a lighting-independent shaded SLAT $Z_s$ based on $I_{in}$, which may contain arbitrary lighting. We then concatenate $Z_s$ with noise of the same shape along the feature channel dimension and feed it into $f_\theta$ to obtain the lighting homogenized-SLAT (LH-SLAT).

% After obtaining the LH-SLAT, we first pre-process $E$ (as described in Sec. \ref{sec:LAD}) and then tokenize it into a lighting condition encoding $C_L$ using the $\mathcal{E}_{l}$ tokenizer. Finally, we use IAD to receive the LH-SLAT, obtaining intrinsic features $h$, and the LAD $\mathcal D_E$ combines the viewing direction encoding $e^v$ and $C_L$ to obtain a 3D Gaussian representation under the corresponding viewing angle and lighting (Eq.~\ref{eq:3dgs}). Through gsplat~\citep{ye2025gsplat}, we render the corresponding relit high dynamic range image $I^{hdr}_{target}$ and use Agx tone mapping\footnote{\url{https://github.com/iamNCJ/simple-ocio}} to map it to low dynamic range as the output result, maintaining consistency with Blender.
Given a single input image $I_{in}$ with unknown lighting and a target high dynamic range (HDR) environment map $E$, our inference pipeline proceeds as follows. Since our method decouples geometry generation from relighting, we first reconstruct a 3D mesh $m$ from $I_{in}$ using Hunyuan3D 2.1 (HY3D 2.1)~\cite{zhao2025hunyuan3d} with default settings. This mesh is then voxelized to provide coordinates for the structurally sparse voxel feature SLAT. 

Following Trellis~\citep{xiang2024structured}, we utilize the pre-trained SLAT flow model $f_s$ to generate an initial shaded SLAT $Z_s$ from $I_{in}$. Note that $Z_s$ inherently contains arbitrary lighting information from the input image. To remove these lighting effects, we concatenate $Z_s$ with noise (matching $Z_s$ in shape) along the channel dimension and feed the result into our fine-tuned corrective model $f_\theta$ to yield the Lighting-Homogenized SLAT (LH-SLAT).

Subsequently, for the target lighting, we pre-process the environment map $E$ (as detailed in Sec.~\ref{sec:LAD}) and encode it into a lighting condition embedding $C_L$ using the Lighting-Aware tokenizer $\mathcal{E}_{l}$. The IAD module then processes the LH-SLAT to extract intrinsic features $h$. Simultaneously, the LAD $\mathcal{D}_E$ integrates the viewing direction encoding $e^v$ and the lighting condition $C_L$ to predict the 3D Gaussian attributes for the specific view and lighting (Eq.~\ref{eq:3dgs}). Finally, we render the relit HDR image $I^{hdr}_{target}$ via gsplat~\citep{ye2025gsplat}. To align the visual output with standard rendering engines like Blender, we apply AgX tone mapping\footnote{\url{https://github.com/iamNCJ/simple-ocio}} to convert the HDR result into a low dynamic range (LDR) image.

\subsection{Network Architectures}
\paragraph{Register Tokens.} Apart from the lighting-aware tokenizer $\mathcal E_l$, and consistent with \citep{xiang2024structured}, our method primarily employs Transformer networks. As depicted in Fig. \ref{fig:pipeline}, the IAD $\mathcal D_I$ comprises 3D shifted window multi-head self-attention (3D-SW-MSA) and a feed-forward network (FFN). Addressing the limitation of the naive 3D-SW-MSA design in \cite{xiang2024structured}, which computes attention solely within local windows and neglects inter-window information exchange, we introduce learnable register tokens. These tokens interact with all windows via 3D multi-head cross-attention (3D-MCA), serving as a global information bridge to facilitate the model's learning of global context. The lighting-aware decoder $\mathcal D_E$ receives intrinsic features $h$, view encoding $e^v$, register tokens, and lighting encoding to generate lighting-dependent features $h^v$. Register tokens and lighting encoding are injected into the network via 3D-MCA. Here, $h$ and $e^v$ are added in a voxel-wise manner to determine which lighting encoding tokens should be attended to under the current viewpoint. The ablation study on the interaction order of viewpoint and lighting information is illustrated in Fig. \ref{fig:ab_arch} and Tab \ref{tab:architectures}.

\paravspace

\paragraph{Loss Functions.} 
For the relightable neural 3DGS synthesis stage, we optimize the model using a composite objective function $\mathcal{L}_{\text{total}}$. This objective is a weighted sum of three primary reconstruction components—HDR reconstruction ($\mathcal{L}_{\text{recon}}$), physically-based material supervision ($\mathcal{L}_{\text{pbr}}$), and shadow-casting ($\mathcal{L}_{\text{shadow}}$)—along with regularization terms for Gaussian primitives:
\begin{equation}
\small
    \mathcal{L}_{\text{total}} = \mathcal{L}_{\text{recon}} + \lambda_{\text{pbr}}\mathcal{L}_{\text{pbr}} + \lambda_{\text{shadow}}\mathcal{L}_{\text{shadow}} + \lambda_{\text{vol}}\mathcal{L}_{\text{vol}} + \lambda_{\alpha}\mathcal{L}_{\alpha}.
\end{equation}
In our experiments, we set the weighting hyperparameters to $\lambda_{\text{pbr}}=0.3$, $\lambda_{\text{shadow}}=0.5$, $\lambda_{\text{vol}}=10,000$, and $\lambda_{\alpha}=0.001$.

\paragraph{Reconstruction Loss ($\mathcal{L}_{\text{recon}}$).} 
We formulate $\mathcal{L}_{\text{recon}}$ to ensure high-fidelity HDR rendering. Before calculating perceptual metrics, we apply AgX tone mapping to both the rendered HDR image $I^{\text{hdr}}_{\text{target}}$ and the ground-truth $I^{\text{hdr}}_{\text{gt}}$, yielding their LDR counterparts $\hat{I}_{\text{target}}$ and $\hat{I}_{\text{gt}}$. The loss combines an L1 distance in the logarithmic domain for HDR consistency, along with SSIM and LPIPS losses on the tonemapped LDR images for perceptual quality:
\begin{equation}
\small
\begin{aligned}
\mathcal{L}_{\text{recon}} = & \quad \mathcal{L}_{1}(\log(I^{\text{hdr}}_{\text{target}} + 1), \log(I^{\text{hdr}}_{\text{gt}} + 1)) \\
& + 0.2 (1 - \text{SSIM}(\hat{I}_{\text{target}}, \hat{I}_{\text{gt}})) \\
& + 0.2 \text{LPIPS}(\hat{I}_{\text{target}}, \hat{I}_{\text{gt}}).
\end{aligned} \label{eq:recon_loss}
\end{equation}

\paragraph{PBR and Shadow Supervision.} 
To guide the model towards physically plausible decomposition, we impose direct constraints on the intermediate PBR feature maps. The material loss $\mathcal{L}_{\text{pbr}}$ supervises the base color ($I^b$), roughness ($I^r$), metallic ($I^m$), and shading ($I^s$) maps against their ground truths:
\begin{equation}
\small
    \mathcal{L}_{\text{pbr}} = \mathcal{L}_{1}(I^b, I^b_{gt}) + \mathcal{L}_{1}(I^r, I^r_{gt}) + \mathcal{L}_{1}(I^m, I^m_{gt}) + \mathcal{L}_{1}(I^s, I^s_{gt}).
    \label{eq:pbr_loss}
\end{equation}
Similarly, $\mathcal{L}_{\text{shadow}}$ employs an $\mathcal{L}_1$ loss to ensure the geometric consistency of cast shadows under novel lighting conditions.

\paragraph{Regularization.} 
To prevent the degeneration of Gaussian primitives (e.g., becoming too large or too opaque) during optimization~\citep{xiang2024structured}, we incorporate a volumetric loss $\mathcal{L}_{\text{vol}}$ and an opacity loss $\mathcal{L}_{\alpha}$:
\begin{equation}
\small
    \begin{aligned}
\mathcal{L}_{\text{vol}} &= \frac{1}{LK}\sum_{i=1}^{L}\sum_{k=1}^K \prod \boldsymbol{s}_i^k + \frac{1}{LK}\sum_{i=1}^{L}\sum_{k=1}^K \prod \boldsymbol{\hat s}_i^k, \\
\mathcal{L}_{\alpha} &= \frac{1}{LK}\sum_{i=1}^{L}\sum_{k=1}^K(1-\alpha_i^k)^2.
    \end{aligned} \label{eq:reg_losses}
\end{equation}
These terms are calculated across the $L$ active voxels, with each voxel predicting $K$ Gaussian primitives. Specifically, $\mathcal{L}_{\text{vol}}$ regularizes the scale components $\boldsymbol{s}$ from the IAD and $\boldsymbol{\hat{s}}$ from the LAD simultaneously.

\begin{figure}[tbp]
    \centering
    \vspace{0pt}    
    \includegraphics[width=0.45\textwidth]{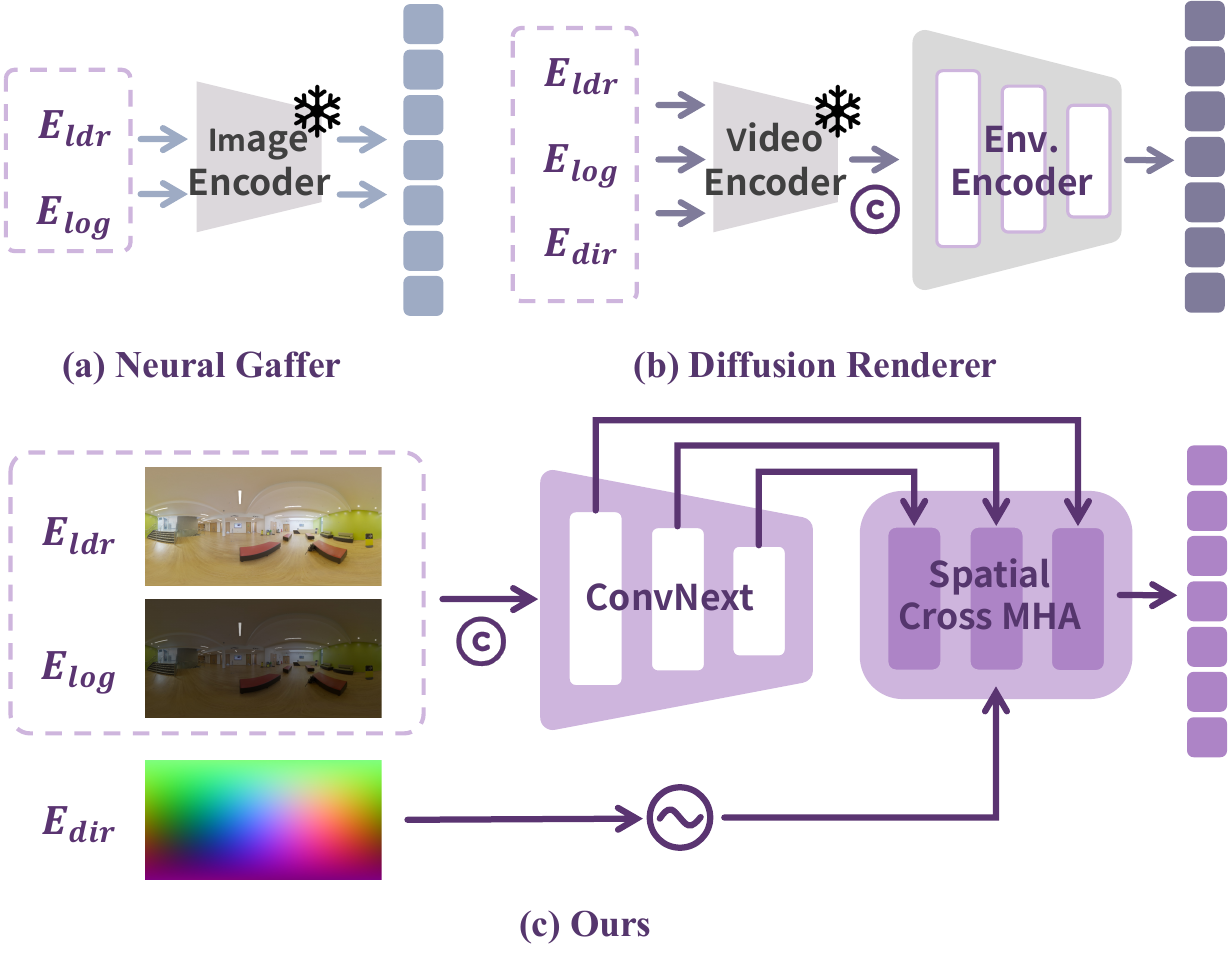}
    \vspace{-5pt}
    \caption{\small Compared to existing HDRI encoding methods, we bind directional information with multi-scale features using positional encoding.}
    \label{fig:hdri}
    \vspace{-12pt}
\end{figure}

\paravspace

\paragraph{Lighting Tokenizer.} As illustrated in Fig. \ref{fig:hdri}, the lighting tokenizer $\mathcal E_l$ is primarily designed to process and inject lighting information into the network for relighting purposes, while also effectively perceiving rotations in the ambient lighting. Similar to Neural Graffer and Diffusion Renderer, as depicted in Fig. \ref{fig:hdri} (a), our approach leverages the $E_{hdr}$ and $E_{log}$ components of the environment map $E$ to provide lighting color characteristics. Neural Graffer encodes the environment map into an image latent space via a pre-trained image VAE, whereas Diffusion Renderer employs a video VAE model to compress $E_{ldr}$, $E_{log}$, and $E_{dir}$ into a video latent space of consecutive frames, thereby accommodating subsequent image or video diffusion model training. 
% However, both of these methods operate in a two-dimensional image space, which poses challenges for direct transference to 3D relighting GS synthesis.

\begin{figure}[t]
\small
\centering
\includegraphics[width=0.95 \linewidth]{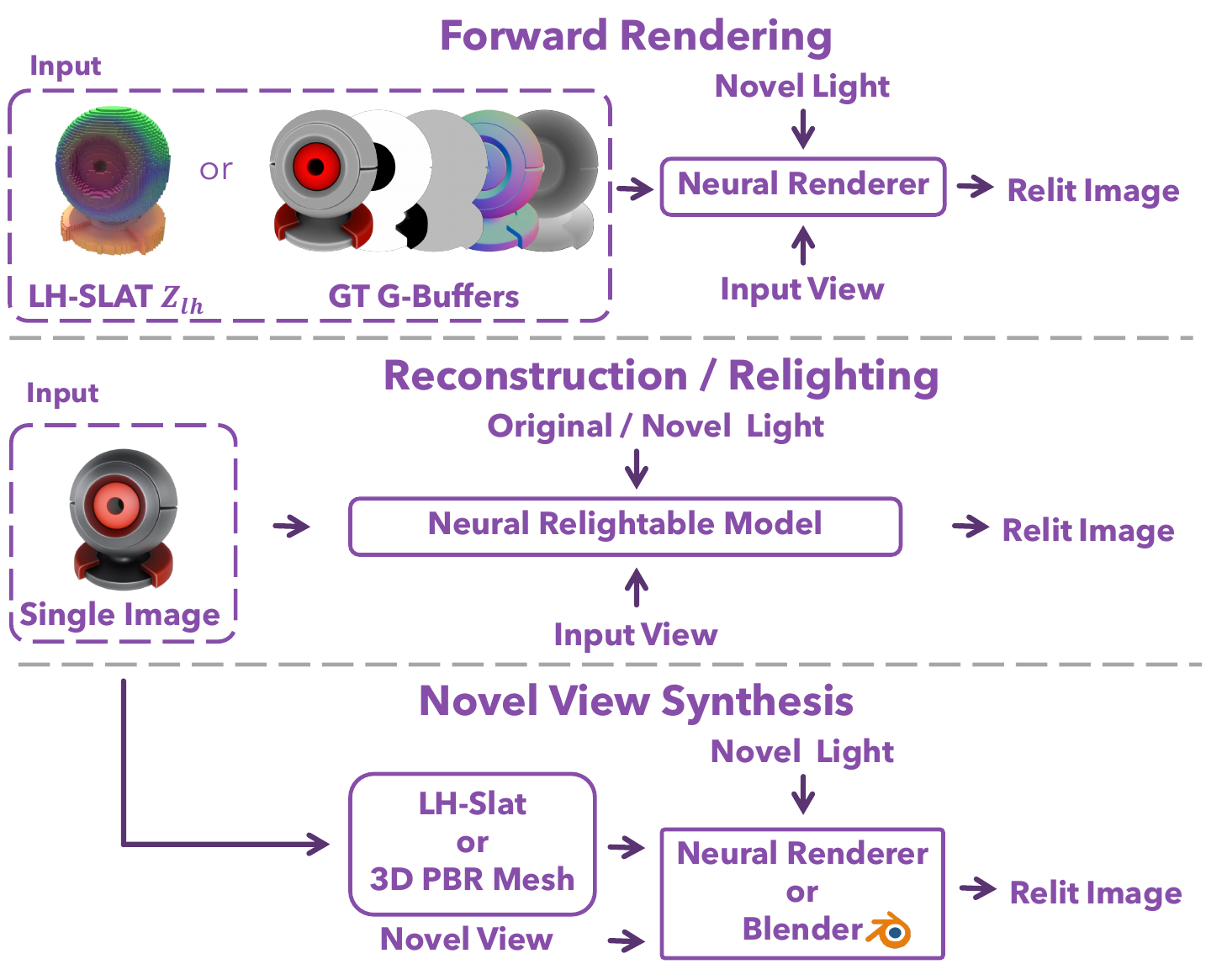}
\vspace{-3mm}
\caption{Schematic illustration of four distinct sub-tasks.}
\label{fig:subtask}
\vspace{-6mm}
\end{figure}

The lighting-aware tokenizer, $\mathcal E_l$, is primarily designed to process and inject lighting information into the network to enable relighting while also effectively perceiving rotations in the environment map. Similar to Neural Graffer and Diffusion Renderer, as depicted in Fig. \ref{fig:hdri}, our approach leverages the $E_{hdr}$ and $E_{log}$ components of environmental illumination to provide lighting color features. Neural Graffer encodes the environment map into the image latent space using a pretrained image VAE's encoder. Diffusion Renderer, in contrast, employs a video VAE model to compress $E_{ldr}$, $E_{log}$, and $E_{dir}$ into a video latent space of consecutive frames, thus accommodating subsequent image or video diffusion model training. 
As shown in Fig. \ref{fig:hdri}(c), the design of $\mathcal E_l$ aims to facilitate the injection of lighting information from the LH-SLAT into relightable 3D Gaussian Splatting (GS). This design addresses two key challenges: First, different materials require sensitivity to varying resolutions of lighting information. For instance, highly rough surfaces require only low-resolution environment maps, whereas high-metallicity surfaces with low roughness necessitate high-resolution maps. To address this, we utilize ConvNext~\citep{liu2022convnet} to extract multi-resolution features from the lighting pyramid and employ a spatial attention mechanism to compute attention scores and exchange information between these resolutions. Second, the model should accurately perceive rotations of the environment map. Neural Graffer requires deforming the environment map itself. Our approach, similar to Diffusion Renderer, can rotate the illumination by adjusting the environment light direction map, $E_{dir}$, requiring only the application of a rotation matrix to the direction vector. However, Diffusion Renderer relies on an additionally trained environment encoder (Env. Encoder). Our method employs a direction-encoding-aware spatial cross-multihead attention (Spatial Cross MHA) to guide visual features at different resolutions using directional information. It combines multi-scale feature fusion to preserve both detailed and global information and utilizes a RoPE+RMSNorm transformer layer for efficient sequence modeling, refer to Fig. \ref{fig:pipeline}. This allows the complex HDRI lighting information to be encoded into conditional tokens suitable for cross-attention, providing high-quality lighting conditions for the renderer. Abstractly, we model the environment map as a set of light source tokens, each encoded with absolute direction vector positional information and multi-scale features. Subsequently, the Lighting-Aware Decoder (LAD), $\mathcal E_l$, can efficiently determine the relevance of each token to the current viewpoint by leveraging viewpoint direction encoding.

% \paravspace

% \paragraph{Datasets.} 

% \section{More Results}

% \subsection{More Comparisons.}
% Additional results for the single-image reconstruction task under random illumination are provided in Fig. \ref{fig:supprec1, fig:supprec2}. For the single-image relighting task with unknown illumination, we present further visual comparisons in Fig. \ref{fig:supprelit1, fig:supprelit2}. Our method recovers more accurate shadows and specular highlights compared to existing 2D diffusion-based relighting models \cite{rgbx,jin2024neural,diffusionrenderer,zeng2024dilightnet}. We also include more qualitative comparisons against 3D generation methods \cite{zhao2025hunyuan3d, chen20243dtopia, sf3d, chen2025meshgen} that produce PBR material maps. Our approach demonstrates more precise material estimation, with highlights and tones that more closely align with the ground truth. To prevent baseline methods from generating non-sensical results, we directly provide a frontal view of the 3D object as input and render the corresponding frontal view for comparison, as shown in Fig. \ref{fig:suppview}.

% \subsection{More Visualization Results.}
% Given a single image and a target environment map, our method enables relightable 3D Gaussian Splatting synthesis and supports rendering multi-view relighted images. We present additional visualizations of relighting results, estimated PBR materials, and shadows in Fig. \ref{fig:supppbr} (as discussed in Sec. \ref{sec:loss}).

\subsection{Experiments Setup}
Fig.~\ref{fig:subtask} shows the quantitative evaluation setup for four sub-tasks—forward rendering, reconstruction, relighting, and novel view synthesis—as summarized in Tab.~\ref{tab:comp}. It illustrates how methods, given different input modalities (e.g., LH-SLAT, G-buffers, single images, or 3D assets), are processed under varying viewpoints and lighting via neural rendering or relightable models to produce relit images.

% \begin{figure}[tbp]
% \small
% \centering
% \includegraphics[width=\linewidth]{figures/recon.pdf}
%   % \captionsetup{font=\smaller}
% \caption{Visual comparison of image reconstruction.}
% \vspace{-2mm}
% \label{fig:recon}
% \end{figure}

% \begin{figure}[tbp]
% \small
% \centering
% \includegraphics[width=\linewidth]{figures/relighting.pdf}
% % \captionsetup{font=\smaller}
% \vspace{-15pt}
% \caption{Visual comparison of relighting.}
% \label{fig:relit}
% \vspace{-3mm}
% \end{figure}

\section{More Results}
\subsection{Additional Comparisons}
\begin{figure*}[tbp]
    \centering
    \vspace{0pt}    
    \includegraphics[width=0.8\textwidth]{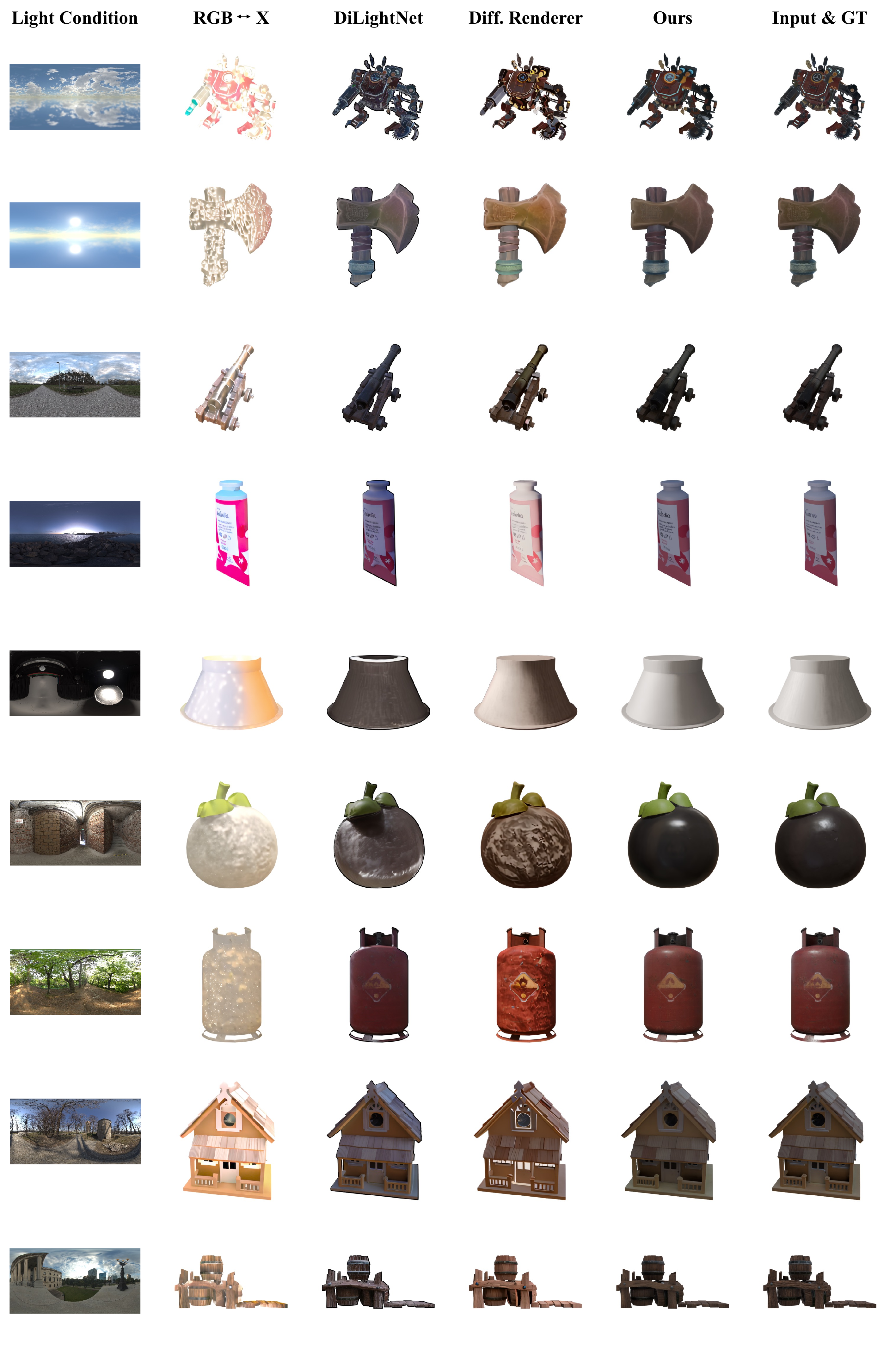}
    \vspace{-5pt}
    \caption{\small Additional qualitative results for single-image reconstruction under random illumination.}
    \label{fig:supprec1}
    \vspace{-12pt}
\end{figure*}

\begin{figure*}[tbp]
    \centering
    \vspace{0pt}    
    \includegraphics[width=0.8\textwidth]{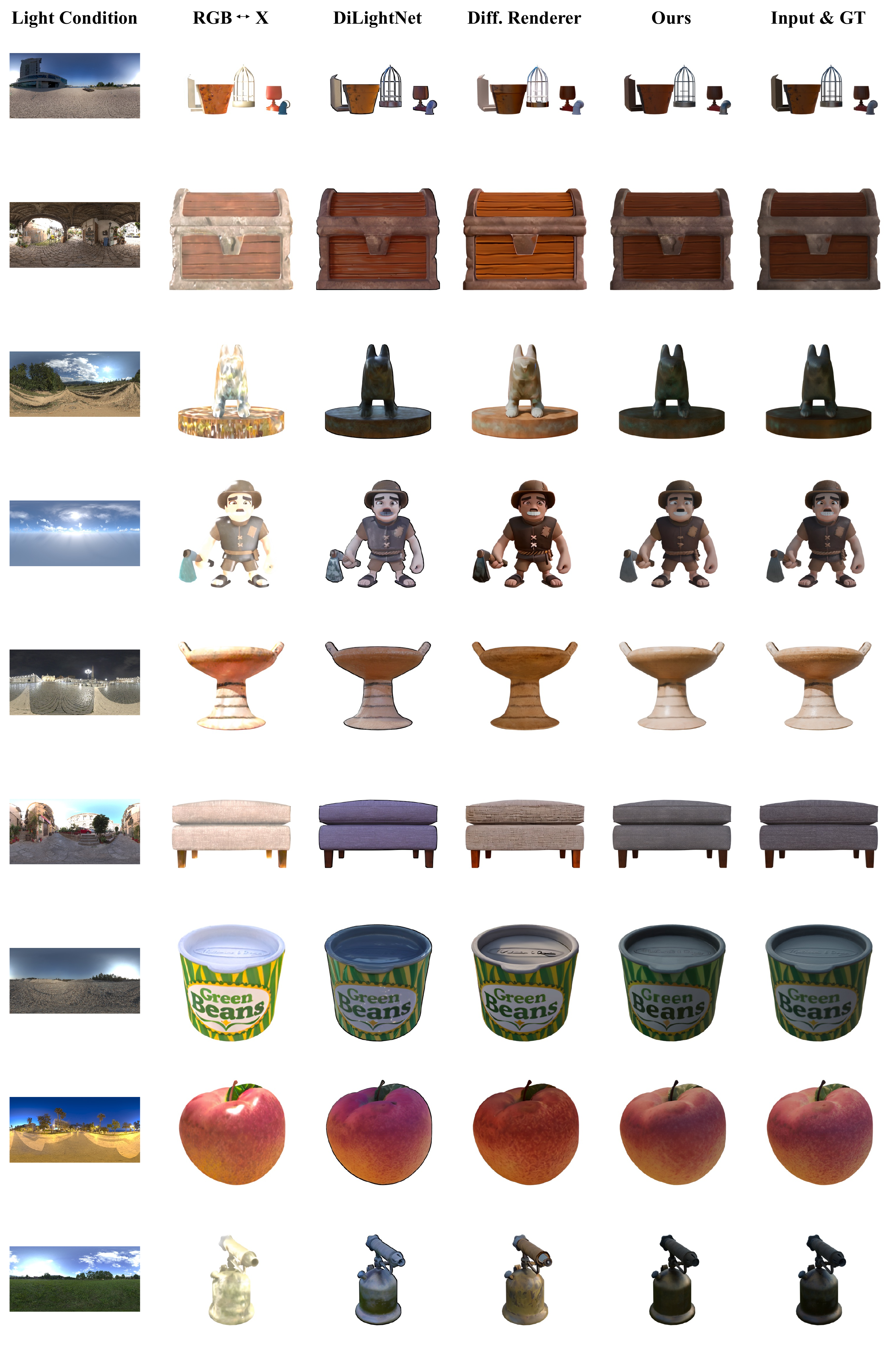}
    \vspace{-5pt}
    \caption{\small Additional qualitative results for single-image reconstruction under random illumination.}
    \label{fig:supprec2}
    \vspace{-12pt}
\end{figure*}

\begin{figure*}[tbp]
    \centering
    \vspace{0pt}    
    \includegraphics[width=0.85\textwidth]{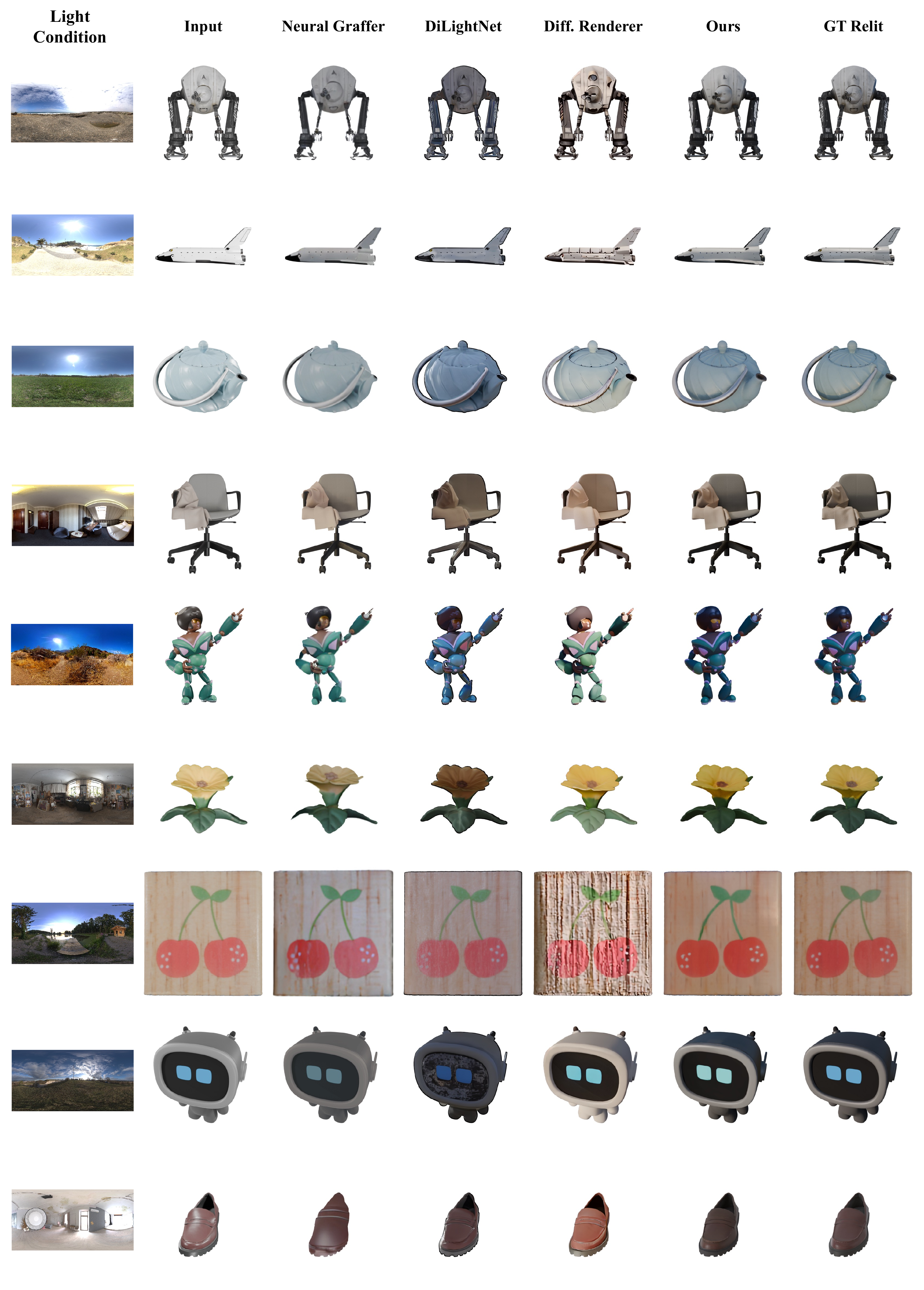}
    \vspace{-5pt}
    \caption{\small More visualization results of relighting and rendering from a single image under unknown illumination.}
    \label{fig:supprelit1}
    \vspace{-12pt}
\end{figure*}

\begin{figure*}[tbp]
    \centering
    \vspace{0pt}    
    \includegraphics[width=0.85\textwidth]{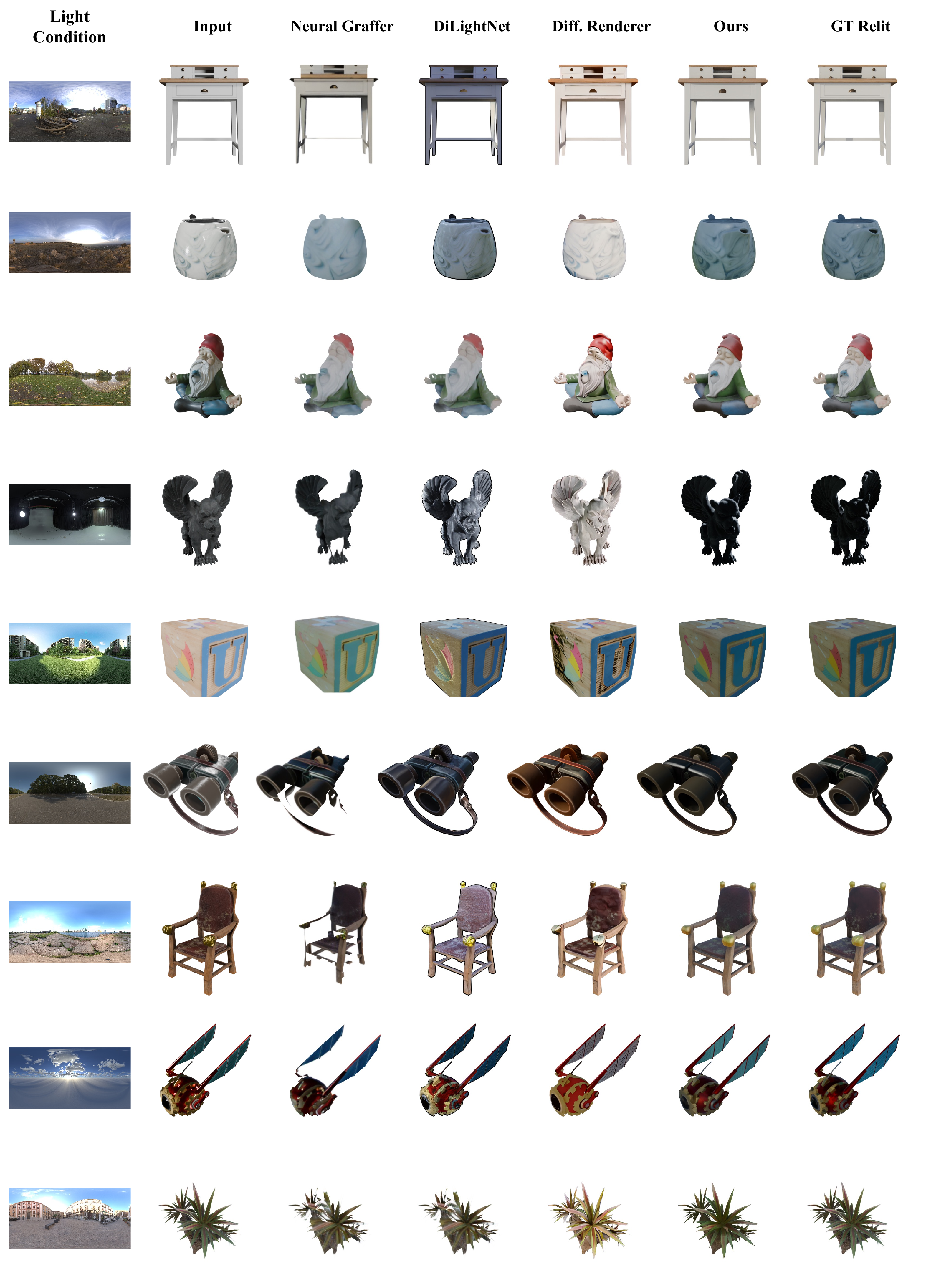}
    \vspace{-5pt}
    \caption{\small More visualization results of relighting and rendering from a single image under unknown illumination.}
    \label{fig:supprelit2}
    \vspace{-12pt}
\end{figure*}

\paragraph{Qualitative Evaluation.}
We provide comprehensive visual comparisons to further substantiate the effectiveness of our method. 
Figures~\ref{fig:supprec1} and \ref{fig:supprec2} illustrate additional results for single-image reconstruction under diverse illumination conditions. 
For single-image relighting with unknown input lighting, we present extended comparisons in Figures~\ref{fig:supprelit1} and \ref{fig:supprelit2}. 
Notably, our method recovers significantly more accurate shadows and specular highlights compared to existing 2D diffusion-based relighting models~\cite{rgbx,jin2024neural,diffusionrenderer,zeng2024dilightnet}, which often struggle with physical consistency.

\paravspace

\begin{figure*}[tbp]
    \centering
    \vspace{0pt}    
    \includegraphics[width=0.9\textwidth]{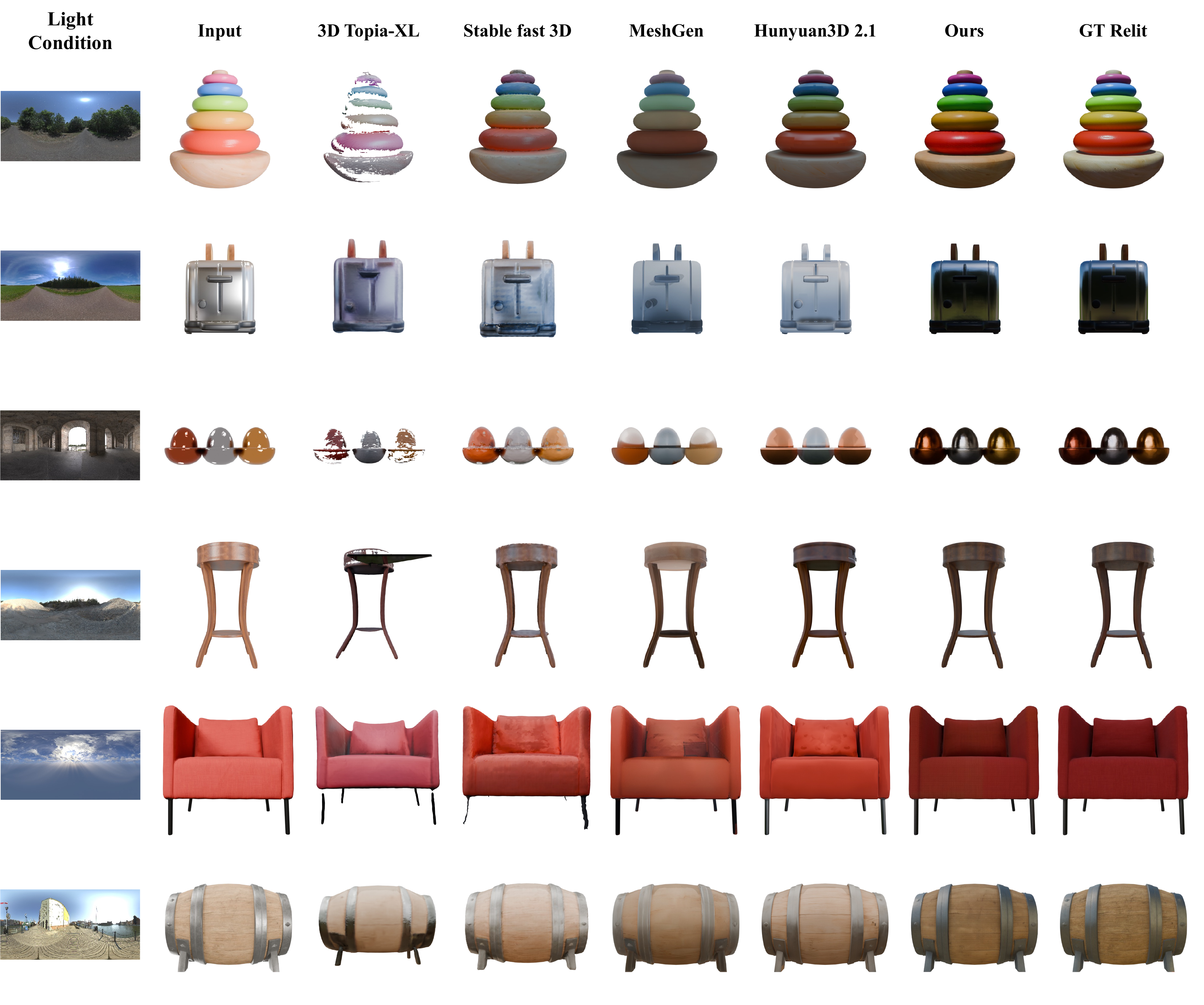}
    \vspace{-5pt}
    \caption{\small Comparison of relighting renderings between our neural rendering method and 3D generation methods that can recover PBR material properties. Our method achieves more stable and accurate rendering results.}
    \label{fig:suppview}
    \vspace{-12pt}
\end{figure*}

\paragraph{Comparison with 3D Generation Baselines.}
We also conduct detailed comparisons against state-of-the-art 3D generation methods capable of producing PBR materials~\cite{zhao2025hunyuan3d, chen20243dtopia, sf3d, chen2025meshgen}. As shown in Fig.~\ref{fig:suppview}, our approach demonstrates superior material disentanglement, yielding highlights and tonal values that align closely with the ground truth. 
To ensure a fair comparison and isolate material quality from geometric failures, we provide the baselines with a fixed frontal view and evaluate the rendered output from the same perspective. This setup mitigates the potential for geometric collapse or severe artifacts in baseline methods, focusing the evaluation on rendering and relighting fidelity.

\subsection{Additional Visualization Results}

\begin{figure*}[tbp]
    \centering
    \vspace{0pt}    
    \includegraphics[width=0.85\textwidth]{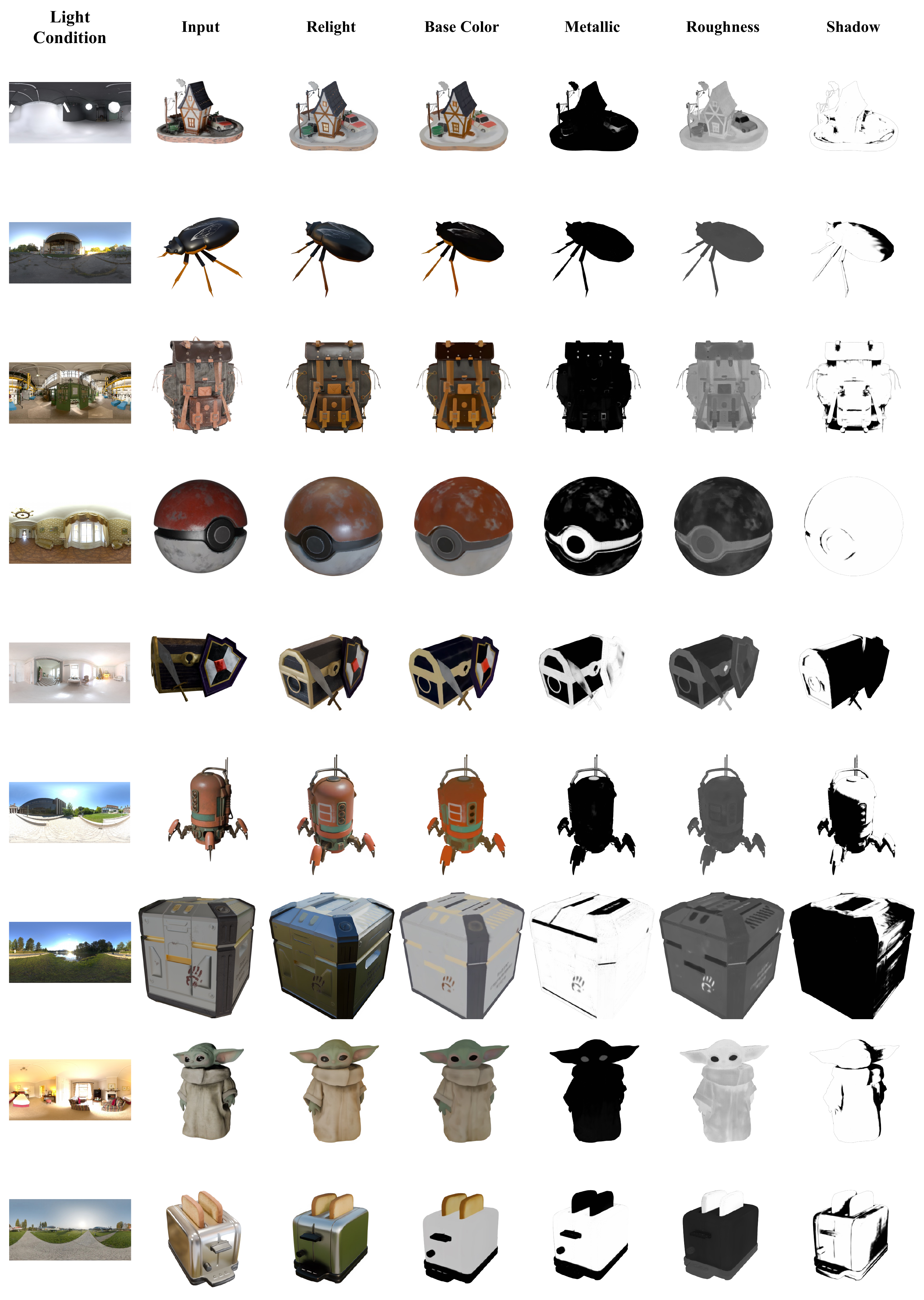}
    \vspace{-5pt}
    \caption{\small Additional relighting results from a single image under target illumination, along with the PBR materials and shadows estimated by our method.}
    \label{fig:supppbr}
    \vspace{-12pt}
\end{figure*}

\paragraph{PBR Material and Shadow Decomposition.}
Leveraging a single input image and a target environment map, our pipeline enables high-fidelity, relightable 3D Gaussian Splatting synthesis with support for multi-view rendering. In Fig.~\ref{fig:supppbr}, we visualize the decomposed PBR material maps (Albedo, Roughness, Metallic) and the generated shadow maps. These visualizations explicitly demonstrate the effectiveness of our physically-based supervision signals (discussed in Sec.~\ref{sec:loss}) in achieving clean and plausible material decomposition.

% \begin{figure}[tbp]
% \small
% \centering
% \includegraphics[width=\linewidth]{figures/pbr.pdf}
% % \captionsetup{font=\smaller}
% \vspace{-15pt}
% \caption{Visual comparison of PBR material estimation with HY3D2.1 and our method.}
% \label{fig:pbr}
% \vspace{-3mm}
% \end{figure}

\section{Discussion}
\label{sec:discussion}

\subsection{Limitations and Future Work}

Despite the robust performance of our framework in generalized single-image relightable 3D Gaussian synthesis, several challenges remain that define coordinates for future research.

\begin{figure}[tbp]
    \centering
    \includegraphics[width=0.5\textwidth]{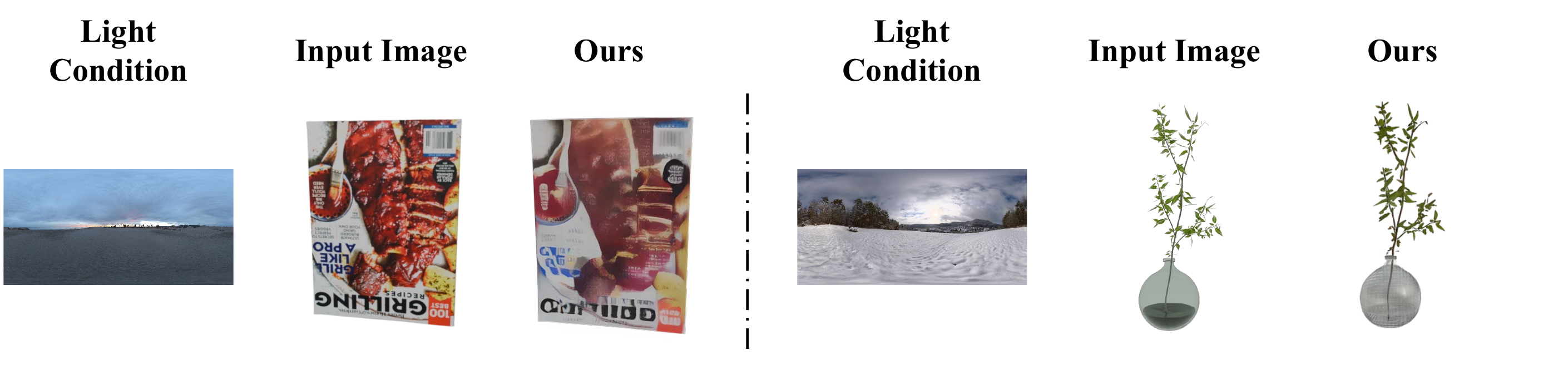}
    \caption{\small \textbf{Failure Cases.} \textbf{Left:} High-frequency details (e.g., text) are blurred due to voxel resolution constraints and VAE compression loss. \textbf{Right:} Transparent objects exhibit checkerboard artifacts. While 3DGS theoretically supports alpha blending, data scarcity in current datasets leads to inaccurate density estimation for refractive surfaces.}
    \label{fig:supplimit}
\end{figure}

\paravspace

\paragraph{Fine-grained Detail Preservation.} 
As illustrated in Fig.~\ref{fig:supplimit} (Left), the reconstruction of high-frequency textures—such as small text—is currently hindered by the feature compression pipeline. The semantic features from DINOv2~\cite{oquab2023dinov2} undergo substantial downsampling to match the resolution of the LH-SLAT voxel grid. This bottleneck inevitably leads to the loss of intricate details. Future iterations will explore multi-scale feature refinement or sparse high-resolution voxel structures to better preserve these fine-grained elements.

\paravspace

\paragraph{Complex Material Modeling.} 
Handling transparent or highly refractive materials remains a significant challenge. While the alpha-blending mechanism of 3DGS natively supports semi-transparency, our model’s ability to represent such surfaces is heavily dependent on the training data distribution. Due to the scarcity of high-quality transparent objects in current large-scale datasets, the model occasionally fails to densify Gaussians sufficiently, resulting in the artifacts shown in Fig.~\ref{fig:supplimit} (Right). Incorporating specialized physics-based transmission losses or curated transparent object datasets could mitigate this issue.

\subsection{Scalability and Generalization}

A core strength of our proposed framework is its inherent scalability across data and architecture, which ensures its long-term viability as a foundation for 3D generative tasks.

\paravspace

\paragraph{Data Scalability.} 
Our modular, multi-stage design allows for independent scaling of different components. Stage 1 (Lighting Homogenization) directly benefits from increasing data volume and diversity, as it learns to suppress complex baked-in illumination—a task that scales effectively with broader data distributions. In contrast, Stage 2 (Lighting-aware Synthesis) is highly efficient due to its feed-forward nature, primarily requiring diversity in material properties and lighting conditions rather than sheer volume.

\paravspace

\paragraph{Architectural Scalability.} 
All core components of our pipeline are based on Transformer architectures, which offer a predictable path for capacity scaling. As demonstrated in our ablation studies (see Tab.~2), increasing model capacity consistently improves rendering quality, particularly for complex specular effects. Furthermore, the LH-SLAT representation is spatially scalable; increasing the voxel resolution allows the model to capture more complex geometries and lighting interactions. This flexibility enables a practical trade-off between inference speed and rendering fidelity depending on the target application.

Overall, the capacity of our model to generalize across diverse object categories and lighting conditions validates our core design philosophy. Our framework validates the effectiveness of jointly designing neural rendering and neural asset stacks, providing a robust and extensible path toward high-fidelity, relightable 3DGS synthesis from arbitrary single images.

\end{document}